\documentclass[nonacm]{acmart}
\usepackage[utf8]{inputenc}
\usepackage{lipsum}
\usepackage{xspace}
\usepackage{graphics}
\usepackage{xspace}
\usepackage{lipsum}
\usepackage{color}
\usepackage{caption}
\usepackage{subcaption}
\usepackage{tikz}
\usepackage{xcolor}
\usepackage{graphicx}
\usepackage{xcolor}
\usepackage{multirow}
\usepackage{array}
\usepackage{etoolbox,enumitem}

\renewcommand\fbox{\fcolorbox{gray!0}{white}}

\begin{document}

\title{TinyML on edX: Widening Access to Applied Machine Learning}
\title{Widening Access to Applied Machine Learning}
\title{TinyML4All: Widening Access to Applied Machine Learning}
\title{TinyML for All: Widening Access to Applied Machine Learning}
\title{Widening Access to Applied Machine Learning: \\TinyML for All}
\title{Widening Access to Applied Machine Learning with TinyML}


\author{
Vijay~Janapa~Reddi\textsuperscript{$\ast$} 
Brian~Plancher\textsuperscript{$\ast$} 
Susan~Kennedy\textsuperscript{$\ast$} 
Laurence~Moroney\textsuperscript{$\dagger$} 
Pete~Warden\textsuperscript{$\dagger$}
Anant~Agarwal\textsuperscript{$\star$$\ddagger$} 
Colby~Banbury\textsuperscript{$\ast$} 
Massimo~Banzi\textsuperscript{$\S$} 
Matthew Bennett\textsuperscript{$\star$} 
Benjamin~Brown\textsuperscript{$\ast$} 
Sharad~Chitlangia\textsuperscript{$\P$} 
Radhika~Ghosal\textsuperscript{$\ast$} 
Sarah Grafman\textsuperscript{$\ast$} 
Rupert~Jaeger\textsuperscript{$\parallel$} 
Srivatsan~Krishnan\textsuperscript{$\ast$} 
Maximilian~Lam\textsuperscript{$\ast$} 
Daniel~Leiker\textsuperscript{$\parallel$}  
Cara Mann\textsuperscript{$\star$} 
Mark~Mazumder\textsuperscript{$\ast$} 
Dominic~Pajak\textsuperscript{$\S$} 
Dhilan~Ramaprasad\textsuperscript{$\ast$} 
J.~Evan~Smith\textsuperscript{$\ast$} 
Matthew~Stewart\textsuperscript{$\ast$} 
Dustin~Tingley\textsuperscript{$\ast$} 
\\[1em]
}

\affiliation{\textsuperscript{$\ast$}Harvard University}
\affiliation{\textsuperscript{$\dagger$}Google\\[1em]
}

\renewcommand{\shortauthors}{}
\raggedbottom
\begin{abstract}
Broadening access to both computational and educational resources is critical to diffusing machine-learning (ML) innovation. However, today, most ML resources and experts are siloed in a few countries and organizations. In this paper, we describe our pedagogical approach to increasing access to applied ML through a massive open online course (MOOC) on Tiny Machine Learning (TinyML). We suggest that TinyML, ML on resource-constrained embedded devices, is an attractive means to widen access because TinyML both leverages low-cost and globally accessible hardware, and encourages the development of complete, self-contained applications, from data collection to deployment. To this end, a collaboration between academia (Harvard University) and industry (Google) produced a four-part MOOC that provides application-oriented instruction on how to develop solutions using TinyML. The series is openly available on the edX MOOC platform, has no prerequisites beyond basic programming, and is designed for learners from a global variety of backgrounds. It introduces pupils to real-world applications, ML algorithms, data-set engineering, and the ethical considerations of these technologies via hands-on programming and deployment of TinyML applications in both the cloud and their own microcontrollers. To facilitate continued learning, community building, and collaboration beyond the courses, we launched a standalone website, a forum, a chat, and an optional course-project competition. We also released the course materials publicly, hoping they will inspire the next generation of ML practitioners and educators and further broaden access to cutting-edge ML technologies.

\end{abstract}


\maketitle

%

\newcommand\blfootnote[1]{%
  \begingroup
  \renewcommand\thefootnote{}\footnote{#1}%
  \addtocounter{footnote}{-1}%
  \endgroup
}
\blfootnote{
\textsuperscript{$\S$}Arduino, %
\textsuperscript{$\P$}BITS Pilani, work done as Harvard intern, %
\textsuperscript{$\parallel$}CreativeClass.ai, %
\textsuperscript{$\star$}edX, %
\textsuperscript{$\ddagger$}MIT %
}

\section{Introduction}

\begin{figure}[t!]
    \centering
\includegraphics[trim={0, 0, 0, -25}, clip, width=\columnwidth]{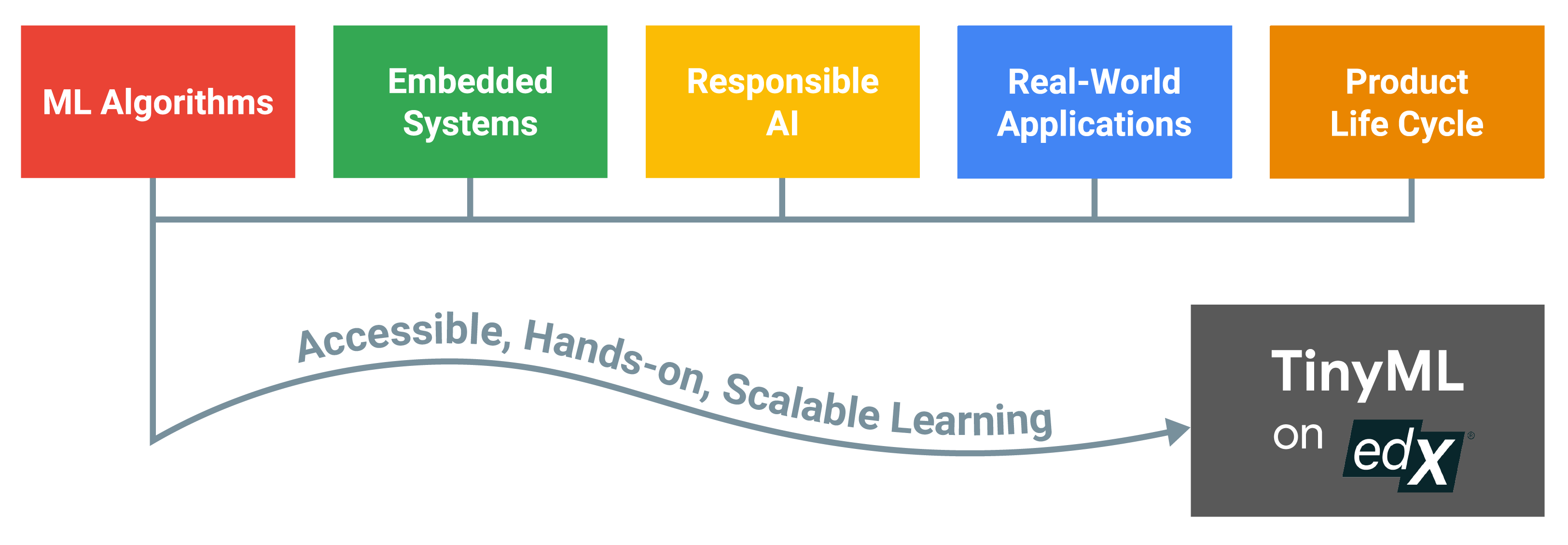}
    \caption{We designed a new applied-ML course motivated by real-world applications, covering not only the software (algorithms) and hardware (embedded systems) but also the product life cycle and responsible AI considerations needed to deploy these applications. To make it globally accessible and scalable, we focused on the emerging TinyML domain and released the course as a MOOC on edX.}
    \label{fig:concept}
\end{figure}

The past two decades have seen machine learning (ML) progress dramatically from a purely academic discipline to a widespread commercial technology that serves a range of sectors. ML allows developers to improve business processes and human productivity through data-driven automation. Given applied ML's ubiquity and success, its commercial use should only increase. Existing ML applications cover a wide spectrum that includes digital assistants~\cite{mitchell1994experience,digital-assistant-2}, autonomous vehicles~\cite{ml-autonomous,ml-autonomous-2}, robotics~\cite{ml-tossing-bot}, health care~\cite{ml-healthcare}, transportation~\cite{ml-transport,ml-transport-2}, security~\cite{ml-security}, and education~\cite{ml-education,ml-education-2}, with new application use cases continuously emerging every few days.

The proliferation of this technology and associated jobs have great potential to improve society and uncover new opportunities for technological innovation, societal prosperity, and individual growth. But it all rests on the assumption that everyone, globally, has unfettered access to ML technologies, which isn't the case. 

Expanding access to applied ML faces three challenges. First is a shortage of ML educators at all levels~\cite{brown_ai_2019,gagne_global_nodate}. Second is insufficient resources, as training and running ML models often requires costly, high-performance hardware, especially as data sets continue to balloon. Third is a growing gap between industry and academia, as even the best academic institutions and research labs struggle to keep pace with industry's rapid progress. Addressing these critical issues requires innovative education and workforce training to prepare the next generation of applied-ML engineers.

This paper presents a pedagogical approach, developed as an academic and industry collaboration led by Harvard University and Google, to address these challenges and thereby increase global access to applied ML. The resulting course, TinyML on edX, focuses not only on teaching the topic by exploring real-world TinyML applications running on low-cost embedded systems, but it also considers the ethical and life-cycle challenges of industrial product development and deployment (see Figure~\ref{fig:concept}).

To improve accessibility we employ both cloud computing and low-cost hardware. We leverage Google's free, open-source TensorFlow and Colaboratory tools along with globally accessible inexpensive embedded devices from Arm and Arduino. We believe hands-on learning that transcends the underlying ML equations is essential. To this end, we focus our approach on TinyML.

\emph{Tiny Machine Learning (TinyML)}, a rapidly growing subfield of applied ML, is a prime candidate for enabling hands-on education globally. This budding area focuses on deploying simple yet powerful models on extremely low-power, low-cost microcontrollers at the network edge. TinyML models require relatively small amounts of data, and their training can employ simple procedures. Furthermore, as TinyML can run on microcontroller development boards with extensive hardware abstraction, such as Arduino products, deploying an application onto hardware is easy. TinyML enables a variety of always-on applications for battery-powered devices--for instance, environmental monitoring and industrial predictive-maintenance analytics. Moreover, the same cost and efficiency benefits open the door to distributed TinyML systems working in concert at the ``edge'' of the cloud-computing network. 

Since TinyML systems are becoming powerful enough for many commercial tasks, learners can acquire skills that can directly apply to their professional careers and future job prospects. The lessons of complete-TinyML-application design, development, deployment and management are also transferable to large-scale ML systems and applications, such as those in data centers and mobile devices. This technology thus provides an attractive entry into applied ML.

Our approach to applied ML through the lens of TinyML provides experience with the complete industrial ML workflow, and it also explores the ethics of software deployment--crucial knowledge for the applied-ML workforce. Creating an ML system is a high-stakes endeavor since inaccurate or unpredictable model performance can erode consumer trust and reduce the chance of success. So understanding ethical reasoning is a crucial skill for ML engineers. To this end, we collaborated with the Harvard Embedded EthiCS program to develop and integrate a responsible-AI curriculum into each course, providing opportunities to practice identifying ethical challenges and thinking through potential solutions to concrete problems, many of which are based on real-world case studies.

To broaden and widen access, we aimed to provide TinyML on a globally available platform that lets users benefit at no cost from instructional resources. We therefore deployed our pedagogical approach on edX, a MOOC provider created by Harvard and MIT that hosts university-level courses in many disciplines. Notably, professionals can choose to prove their newly earned skills with a certificate, available for a small fee, once they satisfy the testing requirements. To foster collaboration and continued learning beyond this edX course, we developed a standalone website, a Discourse forum, a Discord chat, and an optional course-project competition.

We launched the core TinyML edX series, comprising three sequential courses, between October 2020 and March 2021; an optional fourth course is under development. On average, more than 1,000 new students enroll each week. After eight months, over 40,000 have enrolled from 170 countries. They come from diverse backgrounds and experiences, ranging from complete novices to experts who want to master an emerging field. Feedback suggests this strong enrollment may owe to the unique collaborative structure we foster between students, teachers, and industry leaders. Shared ownership between Harvard faculty and staff and Google instructors and engineers appears to give participants confidence they are gaining skills that industry needs both today and tomorrow. Moreover, we recognize that opportunities to interact with experts is both encouraging and validating. 

In summary, our effort to expand access to and participation in applied ML reflects five guiding principles:
\begin{enumerate}
\vspace{.5em}
  \setlength\itemsep{.5em}
    \item Focus on application-based pedagogy that covers all ML aspects. Instead of isolated theory and ML-model training, show how to physically design, develop, deploy, and manage trained ML models. \item Work with industry and academic leaders to aid learners in developing the skills that industry requires today and will require in the foreseeable future.
    \item Raise awareness of the ethical challenges associated with ML and familiarize learners with ethical-reasoning skills to identify and address these challenges.
    \item Prioritize open access to students worldwide by teaching TinyML at a global scale through a MOOC platform using low-cost hardware that is available anywhere.
    \item Build community by providing a variety of platforms so participants can learn collaboratively and showcase their work no matter where they live.
\vspace{.5em}
\end{enumerate}

We hope the approach we devised brings ML to more people. As such we have open sourced our courseware materials which can be found at \href{https://github.com/tinyMLx/courseware}{\texttt{https://github.com/tinyMLx/courseware}}, and note that this paper is part of a broader effort to enable activities such as TinyML4 Developing Countries (TinyML4D) and TinyML4 Science, Technology, Engineering, and Mathematics (TinyML4STEM).


We have organized the rest of the paper as follows. It begins with a discussion of the criteria for increasing access to applied ML (Section~\ref{sec:pedagodgy}). 
We then explain both why TinyML is a useful entry to practical ML and how our courses meet those criteria (Section~\ref{sec:whytiny}). 
Next, the paper describes our series (Section~\ref{sec:courses}) and how we integrated ethics throughout (Section~\ref{sec:ethics}).
We then detail how we quickly and efficiently deployed TinyML by innovating in both multimedia production and the use of MOOCs as well as other online platforms (Section~\ref{sec:atscale}), in addition to
analyzing early data on our courses' impact (Section~\ref{sec:evaluation}). Finally, we introduce the TinyML Open Education Initiative: our effort to further broaden the courses' impact through activities such as TinyML4STEM (Section~\ref{sec:future}). To provide a balanced viewpoint, we discuss some limitations of our approach and suggest alternatives (Section~\ref{sec:limits}). Finally, we conclude the paper with the main takeaways and the lessons learned (Section~\ref{sec:conclusion}).

\section{Challenges and Opportunities} \label{sec:pedagodgy}

We propose three criteria to empower applied-ML practitioners. First, no one size fits all with regard to interest, experience, and motivation, especially when broadening participation. Second, given ML innovation's breakneck pace, academic/industrial collaboration on cutting-edge technologies is paramount. Third, learners who wish to prepare for ML careers need experience with the entire development process from data collection to deployment, and they must understand the ethical implications of their designs before deploying them.

\subsection{Student Background Diversity}


%
%

A major challenge in expanding ML access is that participants begin applied-ML courses with diverse background knowledge, work experience, learning styles, and goals. Hence, we must provide multiple on-ramps to meet the needs of a varied population.

Participants include high-school and university students who want to learn about AI for the first time. Not only will this knowledge empower them to develop cutting-edge applications, but it will also give them an edge in their careers, as many employers expect new hires to have some ML background.

Other participants are industry veterans looking to either pivot their careers toward ML or study the landscape of the  TinyML field. For example, some are computer-systems engineers who want to learn about ML in general. Others are ML engineers and data scientists who want to expand their skills by applying ML. Yet others are doctors, scientists, or environmentalists who are curious about how TinyML technology could transform their fields. 

Other participants are self-taught, makers, tinkerers, and hobbyists who want to build smart ``things'' based on emerging technologies. This group typically operates at the systems level, drawing on prior art, but they want to understand how different components or functional blocks fit together to create intelligent ML devices. 

Given this broad spectrum, we have a unique opportunity to enable inclusive learning for all despite differing backgrounds and expertise. But we must provide multiple on-ramps. Specifically, we chose to structure the course in a spiral that sequentially addresses the same concepts with increasing complexity~\cite{harden1999spiral}. Doing so ensures that not only do participants reinforce fundamentals
while picking up new details, 
but they also master important objectives at every stage. This approach has been shown to improve learning while meeting each individual's objectives~\cite{neumann2017robust}.

\subsection{Need for Academia/Industry Collaboration}



Expanding ML access requires the expertise of academia and industry. Academia is strong in structured teaching: it creates in-depth, rigorous curricula to impart a deep understanding of a field. Conversely, industry is more pragmatic, developing the skills necessary for employment. These approaches are complementary. 

Also, ML is moving rapidly thanks largely to industry's access to rich data. Analysis of ML-research publications at the NeurIPS ML conference suggests industry leads ML innovation~\cite{chuvpilo_2020}. As such, industry has essential domain-specific knowledge that helps ground ML pedagogy in practical skills and real-world applications. 

We belive that academia and industry must work in tandem to deliver high-quality, accessible, foundational, and skills-based ML content. Joining a strong academic institution and a industry leader in technology innovation, with a history of releasing free and accessible resources, makes students confident that they are learning the best skills from the best teachers. 



\subsection{Demand for Full-Stack ML Expertise}

In ML, the ``full stack''\footnote{The term \emph{full-stack} comes from historic career growth in web technologies and the Internet. It began as a series of loosely linked skills but now encompasses web development from the lowest level, the server, to the highest level, the web browser or mobile app.} approach to building and using ML models is the core skill that will define future engineers. The engineers who bring long-term value to their industry are those who have the in-depth knowledge to innovate beyond well-known applications and scenarios. In fact, full-stack developers are now more numerous than all other developers combined, with 55\% of developers identifying as full-stack in a 2020 report~\cite{stack_overflow}.

Our academia and industry collaboration can ensure the course series imparts the full-stack abilities that industry demands. Doing so requires content beyond the narrow, well-lit path of ML-model training, optimization, and inference. We therefore also focus on acquiring and cleansing data, deploying models in hardware, and managing continuous model updates on the basis of field results. Our hope is that learners will gain a whole new set of applied-ML skills and unlock new ideas. 


\section{ML's Future Is Tiny and Bright} \label{sec:whytiny}


We employ Tiny Machine Learning (TinyML), a cutting-edge applied-ML field that brings the potential of ML to low-cost, low-performance, and power-constrained embedded systems and thereby enables hands-on learning. TinyML lets us impart ML-application design, development, deployment, and life-cycle-management skills.

\subsection{Introduction to TinyML}

\begin{figure}[t!]
     \centering
    \subfloat[\label{fig:pico}]{%
         \includegraphics[trim=0 150 0 0, clip, width=.3\linewidth]{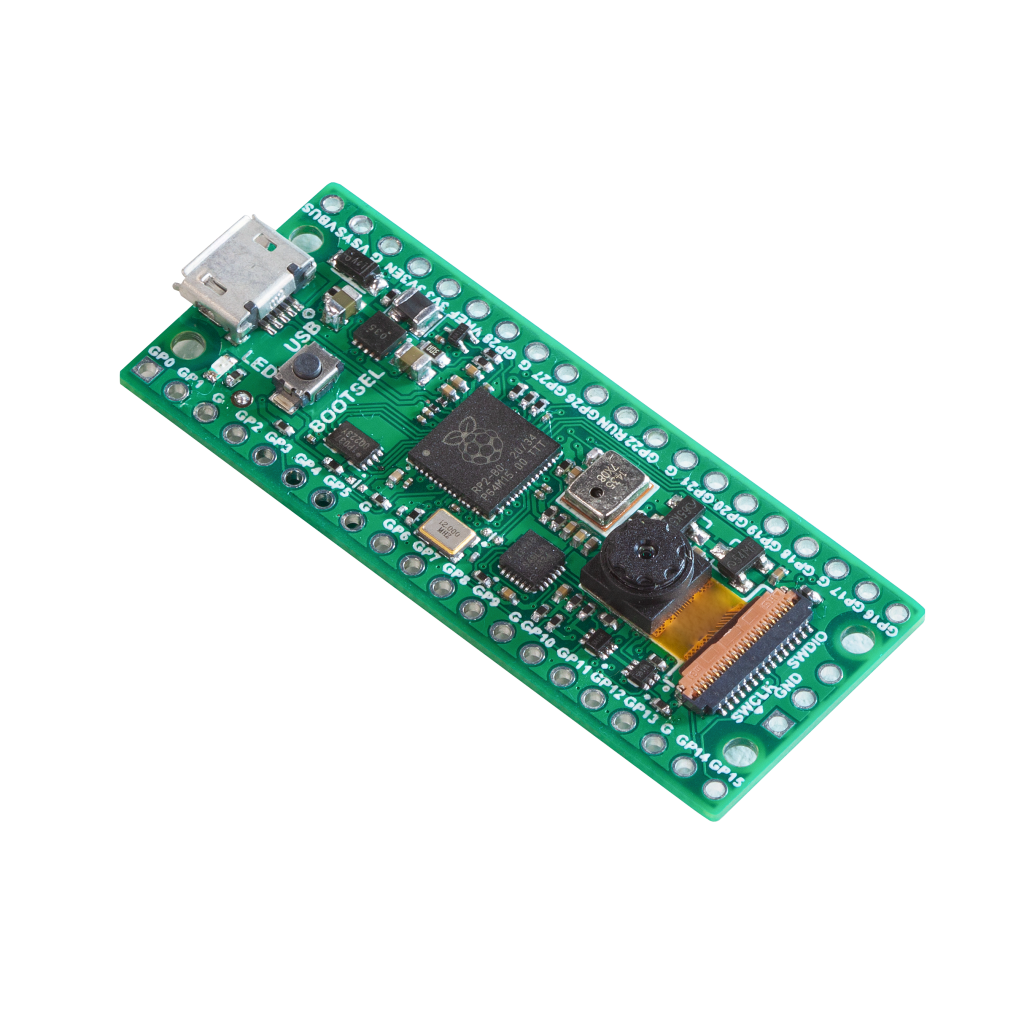}
    }
    \subfloat[\label{fig:arduino}]{%
         \includegraphics[width=.3\linewidth]{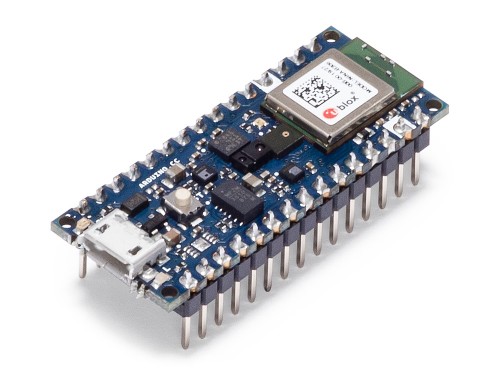}
    }
    \subfloat[\label{fig:stm}]{%
         \includegraphics[width=.3\linewidth]{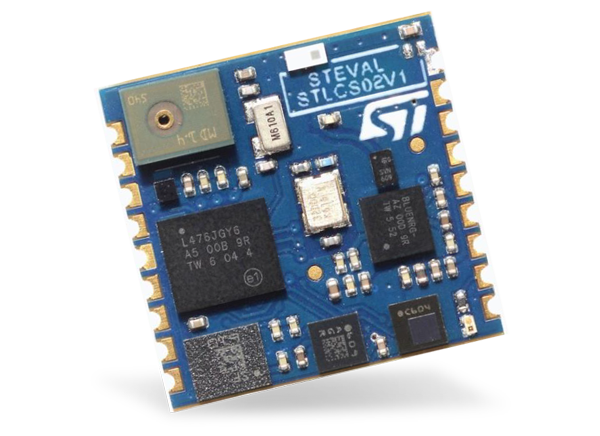}
    }
    \caption{Example TinyML devices: (a) Pico4ML, (b) Arduino Nano 33 BLE Sense, and (c) STMicroelectronics Sensor Tile.}
    \label{fig:devices}
\end{figure}

TinyML refers to the deployment of ML resources on small, resource-constrained devices (Figure~\ref{fig:devices}). It starkly contrasts with traditional ML, which increasingly focuses on large-scale implementations that are often confined to the cloud. TinyML is neither a specific technology nor a method per se, but it acts in many ways as a proto-engineering discipline that combines machine learning, embedded systems, and performance engineering. Similar to how chemical engineering evolved from chemistry and how electrical engineering evolved from electromagnetism, TinyML has evolved from machine learning in cloud and mobile computing systems.

The TinyML approach dispels the barriers of traditional ML, such as the high cost of suitable computing hardware and the availability of data. As Table~\ref{tab:mcu_scale} shows, TinyML systems are nearly two to three orders of magnitude cheaper and more power efficient than traditional ML systems. As such, this approach can serve in embedded devices at little to no cost and can handle tasks that go beyond traditional ML. The TinyML approach also makes it easy to emphasize the importance of responsible AI (Section~\ref{sec:ethics}).

\begin{table}[t!]
\caption{
Cloud \& Mobile ML systems versus TinyML systems.
}
\centering
\scriptsize
\begin{tabular}{l| c | c | c | c | c}
\toprule
Platform                   &  Architecture           &   Memory          &  Storage         & Power         & Price \\
\midrule                        
\textbf{Cloud}           & GPU                     & HBM               & SSD/disk         &               &           \\
E.g., Nvidia V100                & Nvidia Volta            & 16GB              & TB--PB       & 250W          & $\sim$\$9,000    \\
\midrule                    
\textbf{Mobile}          & CPU                     & DRAM              & Flash            &               &           \\
E.g., cellphone                 & Arm Cortex-A78          & 4GB               & 64GB             & $\sim$8W      & $\sim$\$750    \\
\midrule                    
\textbf{Tiny}            &                         &                   &                  &               &        \\
E.g., Arduino Nano               & MCU                     & SRAM              & eFlash           &               &        \\
33 BLE Sense               &  Arm Cortex-M4          & 256KB             &  1MB             & 0.05W         & \$3    \\
\bottomrule
\end{tabular}
\label{tab:mcu_scale}
\end{table}

TinyML supports large-scale, distributed, and local ML tasks. Inference on low-cost embedded devices allows scalability, and their low power consumption enables distribution even to remote locations far from the electric grid. The number of tiny devices in the wild far exceeds the number of traditional cloud and mobile systems~\cite{icinsights}. The ubiquity of tiny embedded devices makes TinyML a candidate for local ML tasks that were once prohibitively expensive, such as distributed sensor networks and predictive maintenance systems in industrial manufacturing settings. 

TinyML applications are broad and continue to expand as the field gains traction. The approach's unique value stems primarily from bringing ML close to the sensor, right where the data stream originates. Therefore, TinyML permits a wide range of new applications that traditional ML cannot deliver because of bandwidth, latency, economics, reliability, and privacy (BLERP) limitations.

Common TinyML applications include keyword spotting, visual wake words, and anomaly detection. Keyword spotting generally refers to identification of words that typically act as part of a cascade architecture to kick-start or control a system, such as a mobile phone responding to voice commands~\cite{ml-kws,ml-kws-mobile}. Visual wake words involve parsing image data to find an individual (human or animal) or object. This task can potentially serve in security systems~\cite{ml-security}, intelligent lighting~\cite{ml-lighting}, wildlife conservation~\cite{ml-wildlife,ml-wildlife-2}, and more. Anomaly detection looks for abnormalities in persistent activities~\cite{chandola2009anomaly}. It has many applications in both consumer and commercial markets, such as checking for abnormal vibrations~\cite{ml-anamoly-vibration} or temperatures~\cite{ml-anamoly-temprature} to provide early warnings of potential failures and to enable preventive maintenance~\cite{ml-anamoly-earlywarning,ml-anamoly-failure}. 

\subsection{TinyML for Applied ML}


An applied-ML engineer should have this full-stack experience to appreciate the impact of the various ML-development stages on the end user. In prototypical ML, such as training large neural-network models in the cloud, learners are unable to participate locally in end-to-end ML development. For example, it is impossible to require them to collect millions of images (akin to ImageNet~\cite{deng2009imagenet}) for large and complex tasks, such as general image classification. Even more difficult is asking all learners to buy the computational resources to train a complex ML model and then evaluate its performance in the real world.\footnote{Just because a trained model performs well on a test data set does not automatically mean it will perform well in the real world.}

By contrast, the small form factor and domain-specific tasks of TinyML enable the full ML workflow, starting from data collection and ending with model deployment on embedded devices. Students thereby gain a unique experience. 
For example, to implement keyword spotting in their native language, course participants learn to collect their own speech data (e.g., by saying ``yá'át'ééh,'' which is Navajo for ``hello''), train a model on that data, deploy it in an embedded device, and test the device in their community.

Such activities create an immersive learning experience, and they are feasible with TinyML because they only require about 30--40 samples of spoken keywords---easy to collect (only from people with their explicit consent) using a laptop with a web browser and microphone. Learners can then train the model using Google's free Colab environment~\cite{bisong2019google} and deploy it in a TinyML device using TensorFlow Lite for Microcontrollers~\cite{david2020tensorflow} or another open-source software technology. This approach allows small keyword-spotting models (about 16KB) to run efficiently on low-cost, highly constrained hardware (less than 256KB of RAM).

\begin{figure}[!tbp]
\centering
    \includegraphics[width=\linewidth]{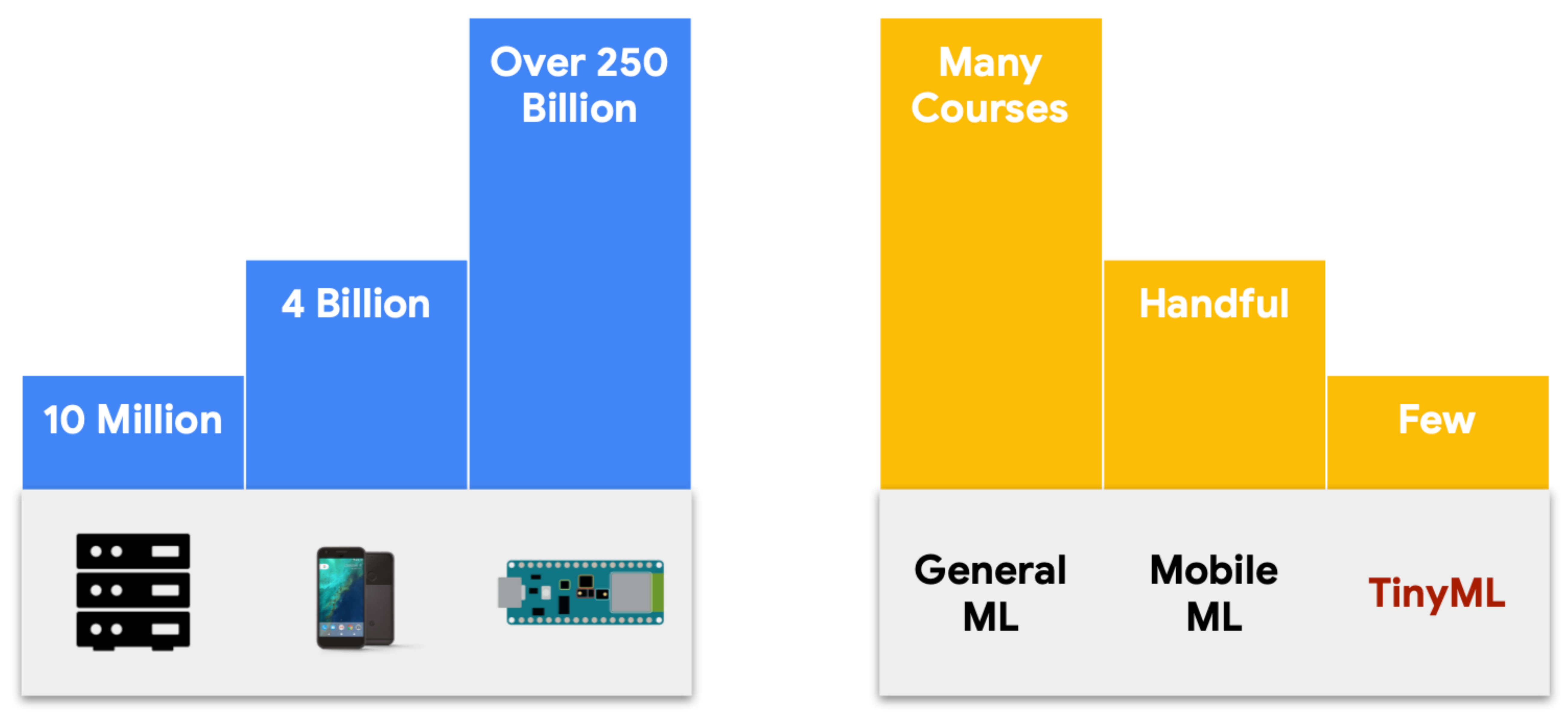}
    \caption{The number of available ML courses is disproportionate to the number of systems in the field.}
    \label{fig:tinymlcount}
\end{figure}

\begin{figure*}[t!]
    \centering
    {\includegraphics[width=\linewidth]{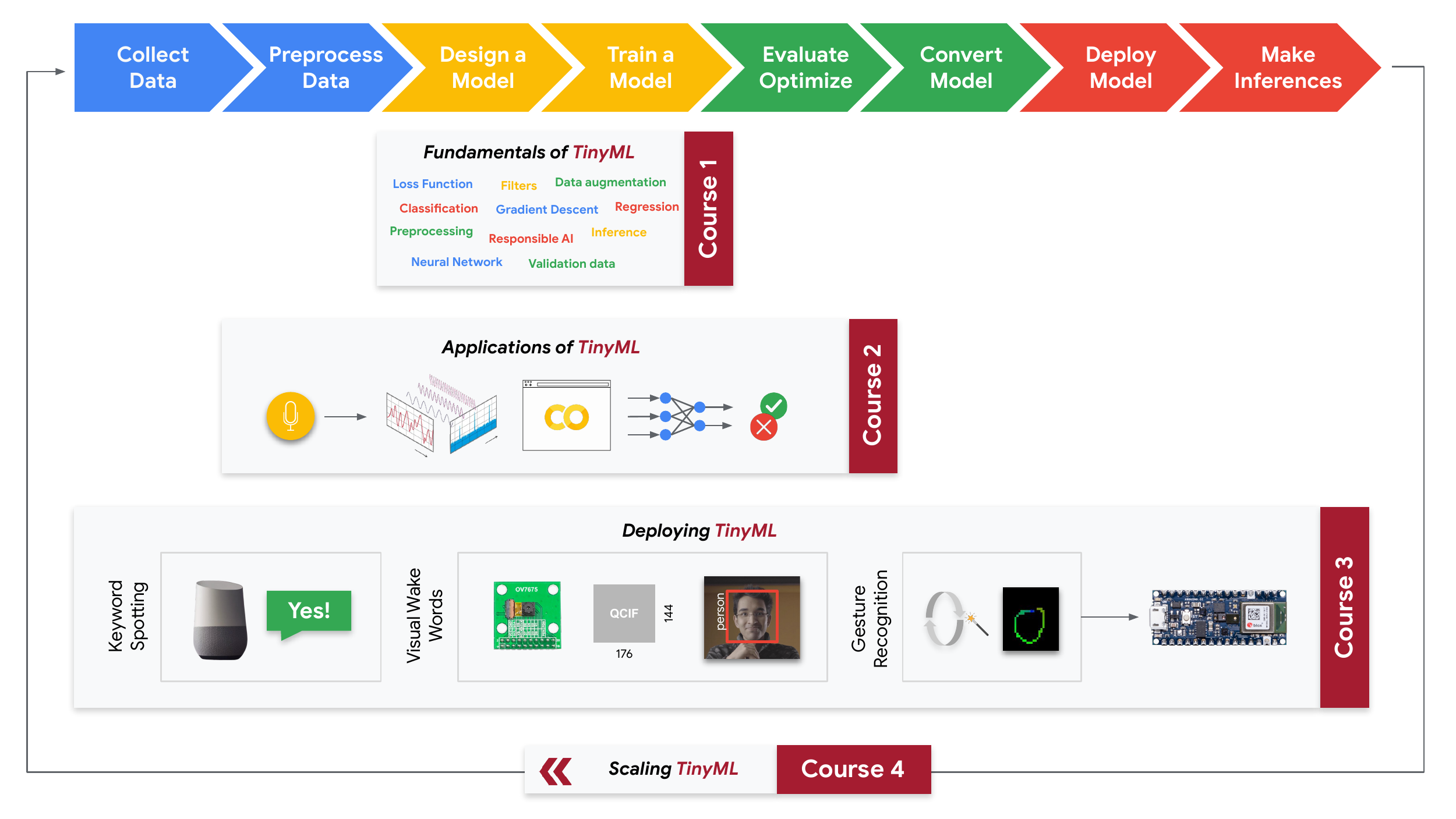}}
    \caption{The ML workflow from data collection to model training to inference. The spiral course design focuses on the neural-network model in Course 1, model application in Course 2, application deployment in Course 3, and, finally, TinyML-model management and scaled deployment in Course 4.}
    \label{fig:workflow}
\end{figure*}

\subsection{TinyML for Expanding Access}
\label{sec:access}

The most difficult task in expanding applied-ML access is making low-cost hardware available anywhere. Cloud-ML technologies cost thousands of dollars, and their physical power, scale, and operational requirements limit their accessibility. Mobile-ML devices are more affordable and pervasive, but their availability is still limited because of network-infrastructure requirements and other factors. 

Research shows that although smartphones have become more affordable, their cost remains a barrier in many low- and middle-income countries (LMICs)~\cite{bahia2020state}. Statista estimates only 59.5\% of the world's population has Internet access, with large offline populations residing in both India and China~\cite{statista_2021}. According to Pew Research, 76\% of individuals in advanced economies have smartphones compared with 45\% in emerging economies. Last in the latter group is India, where only 24\% of the population has a smartphone~\cite{silver_2020}. Students and teachers in many developing countries lack the resources necessary to learn and use traditional ML.

In contrast, TinyML devices are low cost and pervasive. They are readily accessible, enabling hands-on learning anywhere in the world, and their portability eases demonstration of the complete applied-ML workflow in a realistic setting. Furthermore, TinyML applications are more numerous and easier to deploy than mobile-ML and cloud-ML applications. However, despite the wide availability of tiny devices, there is little material for teaching TinyML (see Figure~\ref{fig:tinymlcount}). The number of general-ML courses far exceeds the number of TinyML courses (or, more generally, embedded-ML courses).

\section{Applied-TinyML Specialization} \label{sec:courses}

We developed an applied-ML course specialization focusing on TinyML. Our specialization provides multiple on-ramps to enable a diverse learner population. Moreover, because TinyML is easy to deploy on hardware and test in the real world, it allows us to systematically explore applied ML's vast design space (algorithms, optimization techniques, etc.). It also lets us incorporate responsible AI in all four ML stages: design, development, deployment, and management at scale, which we discuss in greater depth in Section~\ref{sec:ethics}. We hope our description of this applied-ML specialization serves as a roadmap for anyone wishing to adopt the program.

\subsection{A Four-Course Spiral Design}

The TinyML specialization comprises three foundational courses and one advanced course, which we consider optional. Participants would ideally start with the first course and work through the natural progression, but we allow them to go in any order they choose. Depending on their background, they can skip some courses and take the one most relevant to their knowledge and expertise. 


\begin{table*}[t!]
\renewcommand{\labelenumii}{\theenumii}
\renewcommand{\theenumii}{\theenumi.\arabic{enumii}.}
\begin{tabular}{|p{5.55cm} p{5.55cm} p{5.55cm}|}
\hline
& & \\
\small{\textbf{Course 1: Fundamentals of TinyML}}
\begin{enumerate}
    \setlength\itemsep{1pt}
    \setcounter{enumi}{1}
    \item[]
    \begin{enumerate}
        \item Course 1 Overview 
        \item The Future of ML Is Tiny and Bright
        \item Tiny Machine Learning Challenges
        \item Getting Started With ML
        \item The ML Paradigm
        \item The Elements of Deep Learning
        \item Exploring ML Scenarios
        \item Building a Computer-Vision Model
        \item Responsible AI Design
        \item Summary
    \end{enumerate}
\end{enumerate} &
\small{\textbf{Course 2: Applications of TinyML}}
\begin{enumerate}
    \setlength\itemsep{1pt}
    \setcounter{enumi}{2}
    \item[]
    \begin{enumerate}
        \item Course 2 Overview
        \item AI Life Cycle and ML Workflow
        \item ML on Mobile and Edge Devices (Pt. 1)
        \item ML on Mobile and Edge Devices (Pt. 2)
        \item Keyword Spotting (KWS)
        \item Data Engineering
        \item Visual Wake Words (VWW)
        \item Anomaly Detection
        \item Responsible AI Development
        \item Summary
    \end{enumerate}
\end{enumerate} &
\small{\textbf{Course 3: Deploying TinyML}}
\begin{enumerate}
  \setlength\itemsep{1pt}
    \setcounter{enumi}{3}
    \item[]
    \begin{enumerate}
        \item Course 3 Overview
        \item Getting Started
        \item Embedded Hardware and Software
        \item TensorFlow Lite Micro
        \item Deploying Keyword Spotting
        \item KWS Custom-Data-Set Engineering
        \item Deploying Visual Wake Words
        \item Gesturing Magic Wand
        \item Responsible AI Deployment
        \item Summary
    \end{enumerate}
\end{enumerate} \\
\hline
\end{tabular}
\begin{tabular}{|p{5.55cm} p{5.55cm} p{5.55cm}|}
& & \\
\small{\textbf{Course 4: Scaling TinyML}}
\begin{enumerate}
  \setlength\itemsep{1pt}
    \setcounter{enumi}{4}
    \item[]
    \begin{enumerate}
        \item Course 4 Overview
        \item Profiling TinyML Systems
        \item Benchmarking TinyML Systems
        \item Micro NPUs \& Hardware Acceleration
    \end{enumerate}
\end{enumerate} &
\begin{enumerate}
\small
    \setcounter{enumi}{4}
    \item[]
    \begin{enumerate}
      \setlength\itemsep{1pt}
        \setcounter{enumii}{4}
        \vspace{4.5pt}
        \item Neural Architecture Search (NAS)
        \item Machine Learning Operations (MLOps)
        \item TinyML as a Service (TinyMLaaS)
        \item Federated Learning for TinyML
    \end{enumerate}
\end{enumerate} &
\begin{enumerate}
  \setlength\itemsep{1pt}
\small
    \setcounter{enumi}{4}
    \item[]
    \begin{enumerate}
       \setlength\itemsep{1pt}
       \vspace{4.5pt}
        \setcounter{enumii}{8}
        \item Responsible AI Management
        \item Summary
    \end{enumerate}
\end{enumerate} 
\\
\hline
\end{tabular}
\caption{A breakdown of topics in the four TinyML courses. Each one has several activities, including videos, colabs, hands-on labs, quizzes, readings, assignments, tests, and discussion-forum participation. For a detailed overview of the program as well as links to the course materials, visit our courseware Github: \href{https://github.com/tinyMLx/courseware}{\texttt{https://github.com/tinyMLx/courseware}}}.
\label{tab:courseOutline}
\end{table*}

As we mentioned earlier, our application-focused spiral design covers the complete ML workflow, going outward from the middle. The curriculum begins with neural networks for TinyML in Course 1, expands to cover the details of TinyML applications in Course 2, deploys full TinyML applications in Course 3, and application management and scaled deployment in Course 4 (Figure~\ref{fig:workflow}). Our application focus increases learner engagement and enthusiasm~\cite{yang2017instructional}, and our spiral design increases the technical depth over time while reinforcing the main concepts, providing multiple on-ramps and eventually enabling students to create their own TinyML application and deploy it on a physical microcontroller. 

Table~\ref{tab:courseOutline} shows a breakdown of the courses. Roughly, each one takes five or six weeks to complete. For a more detailed and up-to-date overview and links to all course materials, visit our courseware Github at \href{https://github.com/tinyMLx/courseware}{\texttt{https://github.com/tinyMLx/courseware}}.



\subsection{Fundamentals of TinyML (Course 1)}

Course 1 is titled Fundamentals of TinyML. Its objective is to ensure students understand the ``language'' of (tiny) ML so they can dive into future courses. TinyML differs from mainstream (e.g., cloud-based) ML in that it requires not only software expertise but also embedded-hardware expertise. It sits at the intersection of embedded-ML applications, algorithms, hardware, and software, so we cover each of these topics. As Figure~\ref{fig:workflow} shows, the course focuses on a portion of the complete ML workflow. Moving to subsequent courses, we progressively expand participants' understanding of the rest of that workflow. 

The course introduces students to basic concepts of embedded systems (e.g., latency, memory, embedded operating systems, and software libraries) and ML (e.g., gradient descent and convolution). The first portion emphasizes the relevance of embedded systems to TinyML. It describes embedded-system concepts through the lens of TinyML, exploring the memory, latency, and portability tradeoffs of deploying ML models in resource-constrained devices versus deploying them in cloud- and mobile-based systems. 

The second portion goes deeper by focusing on the theory and practice of ML and deep learning, ensuring all students gain the requisite ML knowledge necessary for later courses. Through hands-on coding exercises, students explore central ML concepts, training their own ML models to perform classification using Python and the TensorFlow library in Google's Colaboratory programming environment. 

We provide an overview of embedded systems and ML to ensure students recognize that the topics we cover in the specialization are relevant to their lives and careers, boosting motivation and retention~\cite{dyrberg2019motivational,wladis2014investigation}. For those with sufficient ML and embedded-systems experience, Course 1 is optional. By designing the series with these multiple on-ramps, we can meet participants wherever they are, regardless of their background and expertise.

\subsection{Applications of TinyML (Course 2)}

The objective of the second course is to give learners the opportunity to see practical (tiny) ML applications. Nearly all such applications differ from traditional ML because TinyML is all about real-time processing of time-series data that comes directly from sensors. As Figure~\ref{fig:workflow} shows, we help students understand the complete end-to-end ML workflow by including additional stages, such as data preprocessing and model optimization. Moreover, when we revisit the same stages (e.g., model design and training), we employ spiral design to broach advanced concepts that build on Course 1.

Course 2 examines ML applications in embedded devices. Participants study the code behind common TinyML use cases, such as keyword spotting (e.g., ``OK Google''), in addition to how such front-end, user-facing, technologies integrate with more-complex smartphone functions, such as natural-language processing (NLP). They also examine other industry applications and full-stack topics, including visual wake words, anomaly detection, data-set engineering, and responsible AI.

We take an application-driven approach to teaching the technical components. For example, we use the keyword-spotting (KWS) example to demonstrate the importance of preprocessing sensor inputs, showing the power of FFTs~\cite{fft} and MFCCs~\cite{mfcc} through coding exercises. We additionally explore the importance of holistic architecture by discussing the QoS metrics that evaluate KWS applications and the ``cascade architecture'' (i.e., ML models staged one after another for efficiency) for deploying them~\cite{gruenstein2017cascade}. 
%
As another example, through the lens of the visual wake words (VWW) application, we introduce transfer learning~\cite{transfer-learning}, teaching students to develop their neural-network models without voluminous training data and expensive hardware. Supplementing the theoretical concepts is a coding exercise that employs transfer learning on a pretrained MobileNet~\cite{mobilenet} model to detect whether an individual is wearing a mask---a real-world application that will resonate with learners in light of Covid-19. 
As a final example, we use anomaly detection (AD), in the context of predictive maintenance for manufacturing, to demonstrate the power (and limitations) of supervised learning and deep neural networks by exploring \emph{k} nearest neighbors~\cite{knn}, an unsupervised traditional-ML technique, and comparing it with autoencoders~\cite{baldi2012autoencoders}, an unsupervised neural-network technique. 


The course not only teaches students about TinyML applications and their technical components, but also how to run and test these applications using TensorFlow Lite in Google's Colaboratory programming environment. This step completes the second learning spiral, providing a hands-on opportunity to explore full TinyML applications. The inclusion of TensorFlow Lite lets participants explore important TinyML topics (e.g., neural-network quantization), preparing them to add the next layer in Course 3: physical hardware. 
We intentionally avoided introducing microcontroller hardware until the third course so students could complete the first two entirely for free. They can then make an informed decision on whether to buy the low-cost hardware that Course 3 requires.

\subsection{Deploying TinyML (Course 3)} 

Most instructional ML material focuses on models and algorithms, failing to provide hands-on experience in gathering input or training data, making decisions on the basis of a model's output, and testing models in the real world. Therefore, Course 3 explicitly focuses on demonstrating the complete ML workflow (Figure~\ref{fig:workflow}).

We have found that learners are excited about using the knowledge they gain to solve real problems. But absent guidance in how to build an entire system, many become frustrated because they are unable to apply their knowledge. This issue arises in the form of questions that traditional courses leave out, such as ``How can I find the training data for my problem?'' and ``What threshold should I use to decide whether a classification score is high enough for my application?'' and ``How do I go from an RGB-camera-image byte array to the float-tensor image my model needs?'' 

Providing universally applicable answers to all such questions is impossible. But alerting early learners to them and offering a set of comparable guided practical experiences reduces the frustration before these questions arise in their projects, hobbies, or careers. This is critical, as frustration squelches the desire to master the skills a field requires. Instead, we want to give students the support and confidence they need to overcome challenges and solve problems.


``Deploying TinyML'' mixes computer science and electrical engineering. It gives participants fundamental knowledge and hands-on experiences with ML training, inference, and deployment on embedded devices. Following the spiral pattern, it builds atop many of the techniques and applications from previous courses and adds new technical topics and extensions. Students develop and deploy full applications, such as KWS, person detection, audio/visual reaction, and gesture detection on their own microcontrollers.

\begin{figure*}[t!]
     \centering
     \begin{subfigure}[t]{0.49\textwidth}
         \centering
         \fbox{\includegraphics[trim=0 30 0 30, clip, width=\textwidth]{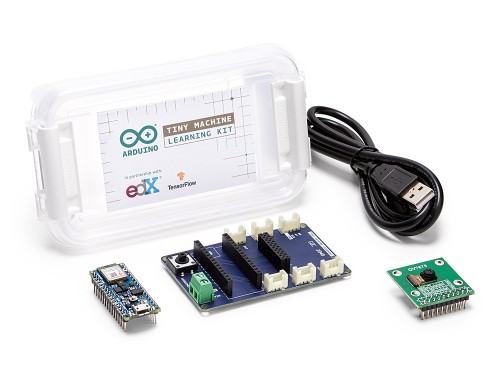}}
     \end{subfigure}
     \hfill
     \begin{subfigure}[t]{0.49\textwidth}
         \centering
         \fbox{\includegraphics[width=\textwidth]{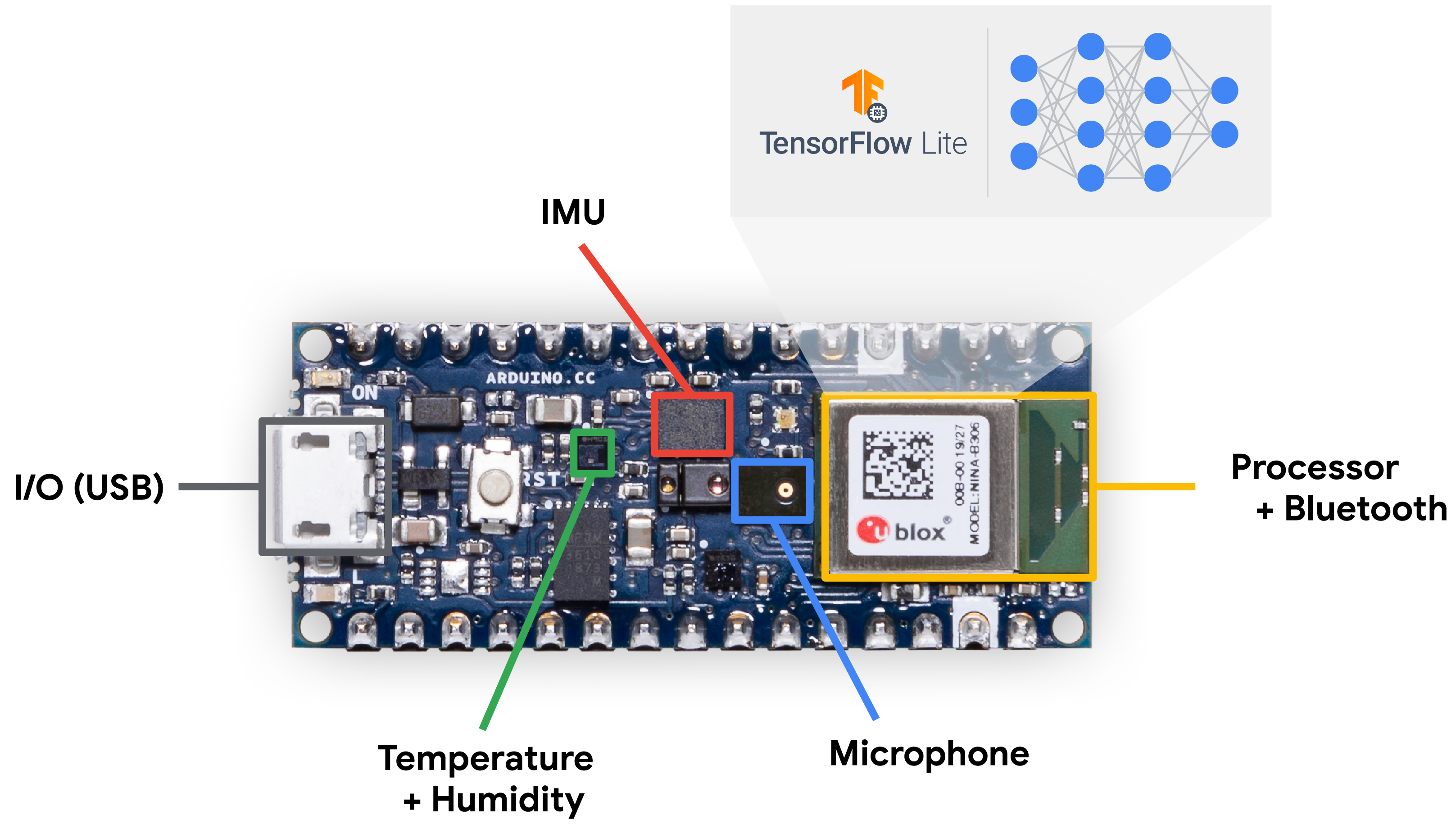}}
     \end{subfigure}
    \caption{Shown at left, the TinyML Kit includes the Arduino Nano 33 BLE Sense~\cite{arduino-nano}, an OV7675 camera module~\cite{ov7675}, and a TinyML shield that simplifies sensor integration. Shown at right, the Nano 33 BLE includes a host of onboard integrated sensors (e.g., temperature/humidity, IMU, and microphone), a Bluetooth Low Energy (BLE) module, and an Arm Cortex-M microcontroller that can run neural-network models using TensorFlow Lite for Microcontrollers.}
    \label{fig:tinyMLkit}
\end{figure*}


The course introduces TensorFlow Lite for Microcontrollers~\cite{david2020tensorflow}, an embedded-ML software library that eases the task of efficiently running ML models on embedded hardware. Students learn how the library works, helping them appreciate the challenges that an embedded-ML-framework engineer faces in the real world. They also examine the library's APIs as they deploy applications such as KWS, VWW, and magic wand to their microcontrollers. 

Building on the concepts from the first two courses, we introduce new concepts such as multitenancy---that is, running more than one ML model at a time---when we revisit certain stages of the ML workflow (Figure~\ref{fig:workflow}). We present the tradeoffs between using multimodal learning that fuses sensor data versus using two separate models to make inferences. The former stresses the first half of the ML workflow (training), while the latter stresses the second half (deployment).

A unique benefit of this course is the exercises that involve the entire ML process. Before they know it, students are implementing an entire TinyML application from beginning to end on a physical device they can hold in their hands. This approach gives our course the unique value of allowing participants to fully develop and use their own TinyML projects at home. This type of hands-on project-based learning is proven to enhance learning, motivation, and retention~\cite{krajcik2006project,vesikivi2020impact}. For instance, participants collect their own custom keyword data for training a KWS model, giving them first-hand experience with the challenges of getting ML models to work accurately. Some find that the \textit{tinyConv}~\cite{tftinyconv} KWS model works well; others find that they must collect more data or adjust the preprocessing. A few individuals in this latter category are perplexed that even those improvements may fail to dramatically increase accuracy, finding that the 16~KB KWS model is too small. The point of the exercises is not necessarily to increase model accuracy, but to instead understand the challenges of applying ML models to the real world.

The course uses the \href{https://store.arduino.cc/usa/tiny-machine-learning-kit}{Tiny Machine Learning Kit}, which we co-designed with Arduino for hands-on, low-cost, accessible, project-based learning. This kit, shown in Figure~\ref{fig:tinyMLkit}, is globally accessible, and includes an Arduino board containing numerous sensors (microphone, temperature, humidity, pressure, vibration, orientation, color, brightness, proximity, gesture, etc.) that enable a wide range of TinyML applications. More importantly, it has a popular Arm Cortex-M-class microcontroller~\cite{martin2016designer} that binds the learning experience to reality. The kit provides everything a student needs to build TinyML applications for image recognition, audio processing, and gesture detection.

After completing Courses 1, 2, and 3, students are eligible to receive the HarvardX/edX Tiny Machine Learning Certificate, testifying they are trained as full-stack TinyML engineers. We offer the certificate because many professional learners desire such awards to enhance their resumes and prove to potential employers that they have mastered particular skills. At this point in the series, participants have not only explored the technical and societal challenges that TinyML poses, but they have also gained hands-on experience with the complete TinyML-application pipeline: collecting data, developing and training models in the cloud using TensorFlow, testing them in the cloud using TensorFlow Lite, and deploying them in hardware with TensorFlow Lite for Microcontrollers.

\subsection{Scaling TinyML (Course 4)} The first three courses bring learners up to speed on designing, developing, and deploying TinyML applications on a device. Course 4 builds on this foundation and considers scaled management of TinyML-application deployments. This advanced course covers two aspects of scaling. The first half focuses on ``scaling up'' the effectiveness of individual TinyML applications through performance benchmarking, model optimization, and hardware/software co-design. The second half focuses on ``scaling out'' TinyML applications from one device to thousands.

In the scaling-up portion of the course, we start by introducing the concept of system-performance profiling through TinyMLPerf~\cite{banbury2020benchmarking} and other open-source, industry-standard benchmarks. Because embedded systems deal with sensor data in real time, the TinyML device must be able to keep up with the data rates. 
In safety-critical systems, such as automobiles, a slow response to new sensor inputs can be life threatening. Knowledge of benchmarking principles is therefore essential; it enables applied-ML engineers to compare ML systems in a fair and useful manner and to make informed decisions when selecting a device for a particular task.

Next, given a suitable system, we discuss the art of picking a suitable model, emphasizing when and whether to be more code-centric (improve the model) or data-centric (improve the data set). The most accessible means of decreasing model latency is to change the neural-network architecture. Students can accelerate inference without changing any code by designing a new model that is sufficiently accurate yet requires fewer calculations. But designing ML-model architectures is difficult and time consuming because of the many decisions that affect model quality and latency: what type of neural network to choose, what size to make it, how many hidden layers and neurons to include, how to best initialize the network, and so forth. Fortunately, services have emerged to help design network architectures automatically. Cloud services such as AutoML~\cite{CloudAut6:online} and techniques such as neural-architecture search (NAS)~\cite{zoph2018learning} allow even developers with limited ML expertise to quickly train high-quality models that meet their needs. In TinyML, such services are essential because achieving efficiency means co-designing the MCU hardware, software, and models, a challenging task for humans. The course therefore introduces concepts such as AutoML and NAS, explaining the fundamentals so students can employ high-level automation tools with confidence.

\begin{figure}[t]
    \centering
    \includegraphics[width=\columnwidth]{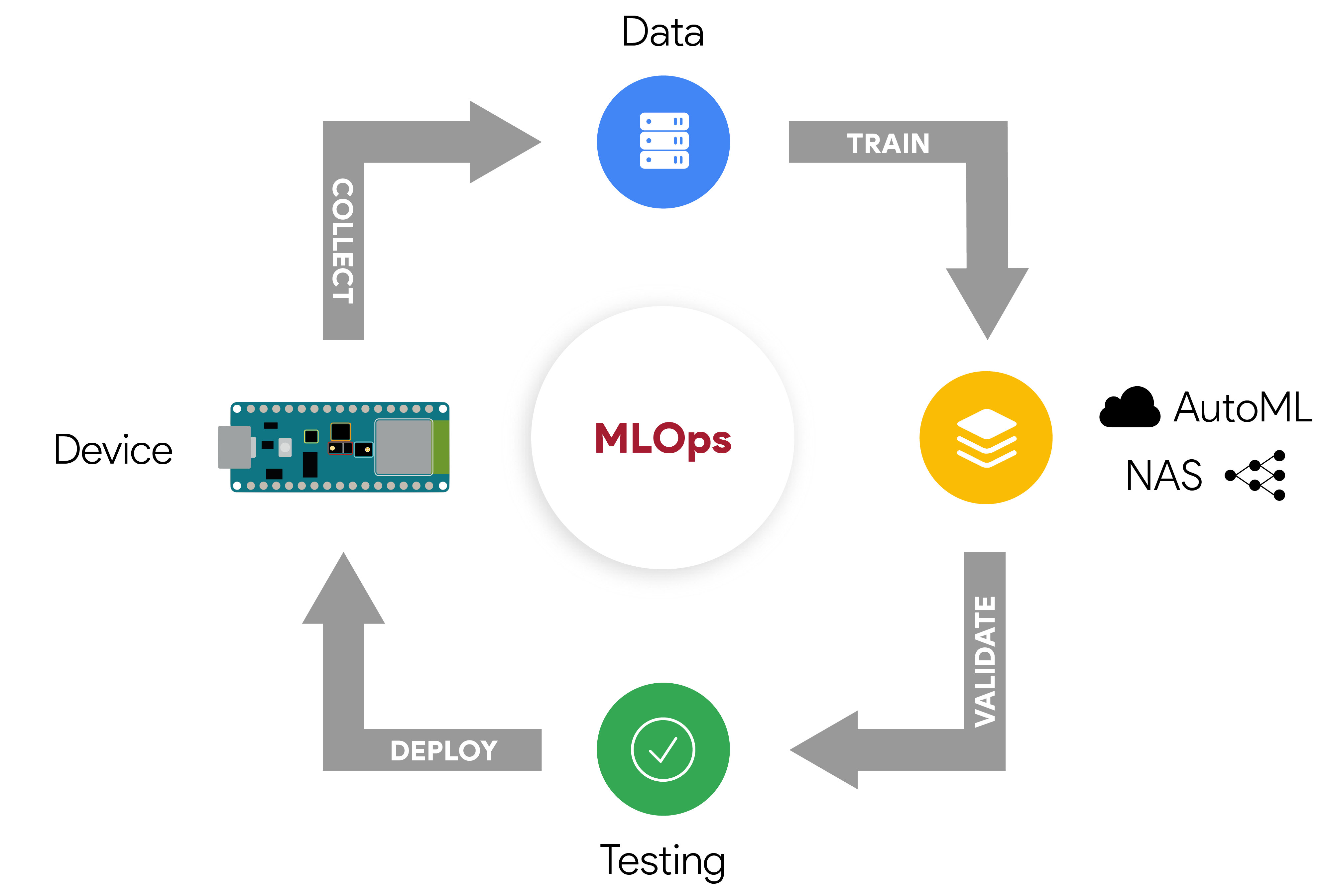}
    \caption{Scaling TinyML through MLOps.}
    \label{fig:mlops}
\end{figure}

The second half of the course focuses on ``scaling out'' TinyML applications from one device to thousands. Applied-ML engineers must know how to manage such deployments as a production ecosystem may involve hundreds or thousands of devices. We thus offer a preview through the TinyML lens. We start by leveraging ``ML operations'' (MLOps) to develop, monitor, and improve a TinyML application. MLOps automates the complete workflow (Figure~\ref{fig:workflow}) as Figure~\ref{fig:mlops} shows. We discuss ways to automatically manage and process data, train ML models, version them, and evaluate, compare, and deploy them, all from the perspective of complete MLOps platforms. This approach introduces the advantages of an automatic ML workflow, which include managing the overwhelming complexity of ML deployments, reducing the burden of maintaining in-house ML knowledge, being more scientific, easing long-term maintenance, and ultimately improving the model's performance in the field.

In addition, we also introduce TinyML as a service (TinyMLaaS), which allows production ecosystems to easily manage and integrate heterogeneous TinyML devices. Because embedded TinyML devices are specialized---from the ML compilers to the operating system and ML hardware---to achieve ultra-low-power energy efficiency, a highly fragmented ecosystem can result. Fragmentation limits a precompiled ML inference model's portability when the hardware changes. The model then requires recompilation for a particular device, leading to deployment complexity. To reap the efficiency benefits of hardware heterogeneity while coping with the fragmented ecosystem, we need a new ``as-a-service'' abstraction. To this end, we introduce learners to TinyMLaaS~\cite{doyu2021tinymlaas}, a general method for tailoring an ML inference model to a specific device. It is a software abstraction layer that gathers information about the target device---such as the CPU type, RAM and ROM sizes, available peripherals, and underlying software---to generate the correct compiled inference model. The designated device then automatically downloads this generated inference model, and the process repeats for all other devices. Because TinyMLaaS enables firmware-over-the-air (FOTA) and software-over-the-air (SOTA) updates, it introduces privacy and security concerns for both the model and the data. Hence, we also cover how large device networks can train models while maintaining user privacy through federated learning~\cite{konevcny2016federated}.

Participants who complete the four courses will have learned all the fundamentals of ML-model design, development, deployment, and management through the TinyML lens. This knowledge is invaluable for career advancement in this quickly emerging field.



\subsection{Student Activities}

\begin{figure}[t!]
    \centering
    \includegraphics[width=\columnwidth]{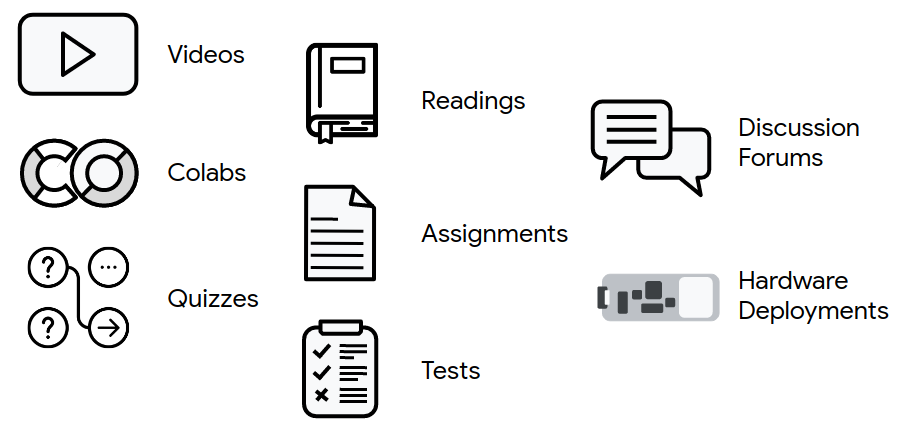}
    \caption{We employ a wide array of learning activities to give students an immersive, self-paced online experience.}
    \label{fig:courseActivities}
\end{figure}

People learn differently~\cite{pashler2008learning}. To support many learning styles, we implemented proven strategies~\cite{lockman2020online} and a variety of methods (Figure~\ref{fig:courseActivities}). Our approach mixes video lectures, short readings, and coding exercises in Google's Colaboratory programming environment to teach and reinforce the course's main technical components. Thus, visual, auditory, and experiential participants all learn by their preferred method. 

We keep the videos short (4--10 minutes). Research shows that people learn better from numerous short content modules than from a few long ones~\cite{guo2014video}. Because some students will find that many of the coding components are new, we provide walk-throughs of major sections, numerous comments describing the code, and introductory text to explain the purpose of each code snippet.

In the first two courses, each section builds toward a coding assignment in Colaboratory to encourage project-based exploration and creativity. The assignments in Course 3 expand to full-on hardware deployment that lets students hold their own designed, trained, and deployed model in their hands and test it in the real world. 

The activities grow in complexity and detail as students progress through the courses, following our spiral-design principle. Students thus gain confidence throughout, as complete application deployments can be challenging. Finally, we sought the input of industry veterans on our course staff to ensure the hands-on activities build relevant full-stack skills.

The courses also include formative multiple-choice quizzes throughout, focusing on the main concepts so students see their progress even if they do not understand every line of code. The quizzes also reinforce the importance of high-level tradeoffs and applied-ML concepts, which will be relevant to ML careers even if the technical stack changes. For those pursuing a paid certificate, we also included summary tests at the end of each section.

Finally, we provide many discussion forums that allow students to ask questions and get answers. Forums allow the course staff to support all participants regardless of their location. They also serve the dual function of building a community around the course. Our forums encourage students to ask any and all questions and to answer them for one another.  

Each activity includes a strong ethics component, which we describe in detail later (Section~\ref{sec:ethics}). Briefly, however, we ask many open-ended questions to elicit student opinions on the opportunities and challenges of responsible TinyML-application design, development, and deployment. As the literature predicts~\cite{lockman2020online}, many of these questions have led to conversations and debates between our online participants, despite their different geographic locations, ages, and technical backgrounds.


\subsection{Accessible, Hands-on Learning}\label{sec:kit}

To enable hands-on learning anywhere in the world, we need a low-cost, self-contained yet extensible, approachable yet representative, flexibly abstracted, and globally accessible TinyML platform. Once again, microcontrollers are promising because they are inexpensive and widely available. So, to provide an easily accessible out-of-the-box experience, we custom designed the Tiny Machine Learning Kit (Figure~\ref{fig:tinyMLkit}) with Arduino. This section describes the kit and its development.


\textbf{Systematic selection.} The range of TinyML hardware and software options is wide, but we believe an ideal solution is fully self-contained yet extensible, approachable yet representative, and flexibly abstracted. As such, we searched for one that not only made it simple to integrate the sensors required for the course but also supported easy integration of additional sensors for future study.

Putting application-level software aside leaves two fundamental elements: 
hardware, from the microprocessor to peripherals and discrete circuitry, and software, from the application layer (our focus) to the silicon layer. An initial constraint on the field of potential microprocessors is the need to support TensorFlow Lite for Microcontrollers~\cite{david2020tensorflow}, which is written in C++ and requires that the microcontroller support 32-bit computing.

We developed criteria for compatible microcontroller development boards, recognizing that
an integrated off-the-shelf product would greatly increase accessibility. These criteria include a small form factor (it is \emph{tiny} ML, after all), a low power budget (efficiency is critical to edge computing), a small system memory (some controllers have large memories, making them less accessible and limiting their range of application), sufficient clock speeds, wireless-communication capability (to enable periodic reporting and/or distributed systems), select sensor integration, and serial channels for extensibility. We defined similar criteria for the accompanying software, comprising the development environment, embedded framework, and logistics (fast, reliable distribution). Next, we added weights to the selection criteria and compiled the candidates in a Pugh matrix~\cite{pugh}. We ranked a field of about two dozen hardware products, giving some preference to controllers that had undergone more-extensive testing---in particular, Arm's Cortex-M series~\cite{martin2016designer} and Espressif Systems' products (namely, the ESP32t)~\cite{esp32}. Both of these embedded systems are widely popular.

\textbf{The TinyML kit.} 
Ultimately we selected the Arduino Nano 33 BLE Sense~\cite{arduino-nano} because it uniquely blends expert embedded-systems engineering and remarkable isolation of the application developer from many low-level hardware details~\cite{jamieson2011arduino}. Furthermore, the Arduino framework and its software APIs (``cores'') fit naturally with our spiral design. Arduino's many libraries and simple IDE are easy for inexperienced students to learn, yet it typically permits those interested in the ``bare metal'' to work their way down the embedded-software stack. Moreover, the Nordic nRF52840 Cortex-M4-based controller~\cite{nordic} on the Nano 33 BLE Sense development board, along with its Mbed real-time OS~\cite{mbed}, represent industry-level hardware and software.

We also developed the Tiny Machine Learning Shield to enable plug-and-play integration of sensors that the Nano 33 BLE Sense lacks. In particular, it eliminates the need for users to make 18 individual connections between the microcontroller and the low-cost camera module we selected for the course---the OV7675~\cite{ov7675}, which typically sells for about US\$2. A series of Grove connectors~\cite{grove} line each side of the shield for connection to numerous additional sensors, which students can purchase for their own projects and integrate without soldering or low-level circuit design. 

We bundled the Nano 33 BLE Sense with the shield, the OV7675 camera module, and a USB cable to form the Tiny Machine Learning Kit (Figure~\ref{fig:tinyMLkit}); learners can purchase a single item and be fully prepared for Course 3 for US\$49.99. To accommodate those few who prefer to purchase elements individually, we provide wiring diagrams and a custom Arduino software library so they can readily swap the OV7675 for the related OV7670 camera module.

\textbf{Alternatives.} In the months after we developed and announced our TinyML kit, similar boards emerged to provide alternative options. For example, the Pico4ML by ArduCam~\cite{pico4ml} is a notable single-board example that comes complete with a microphone, inertial measurement unit (IMU), and camera module, and is suitable for the course exercises.
We are working to support some of these new and exciting hardware platforms to give students more flexibility with their projects.
\section{Ethical \& Responsible AI} ~\label{sec:ethics}

Ethical and Responsible AI is about putting people, social benefit, and safety first. More specifically, ethical AI emphasizes the need for ML engineers to safeguard user privacy and security, mitigate algorithmic bias and discrimination, and ensure ML models perform reliably after deployment. It also extends to developing consumer trust. In this section, our goal is to shift learners from thinking about which ML technology is feasible to which is useful, with an understanding of how it will influence users and society.

\subsection{Ethical Consideration of Ubiquitous ML}

TinyML offers many helpful features, ranging from data privacy and security to low latency and high availability. Coupled with low-cost embedded hardware, these features make it a pervasive technology that can enable ML everywhere. TinyML sensors will monitor the environment in which they are deployed, be it mechanical or human, around the clock. With the prospect of ML everywhere comes a pressing need to address privacy, drift, bias, and other ethical issues. 

Fortunately, TinyML allows us to incorporate responsible AI into all four ML stages: design (Course 1), development (Course 2), deployment (Course 3), and scaling (Course 4). By embedding ethics into each TinyML course, we communicate the technology's ethical and social dimensions in a personal and practical manner. 

To achieve deep integration, we follow the Embedded EthiCS pedagogy at Harvard~\cite{EmbEth}, where philosophers participate directly in computer-science courses to teach students how to think through the ethical and social implications of their work. We collaborated with a philosopher from this program to co-develop and include such material in our curriculum. Her commitment to learning the technical aspects of TinyML enabled us to customize the ethical content to meet the unique course needs of TinyML.

By distributing responsible AI throughout the series, covering the entire ML workflow, students discover how ethical issues permeate all aspects of their work. Our aim is to introduce them to the conceptual tools for navigating these issues, in hopes they will view responsible AI as an active enterprise. Next, we describe our pedagogical goals for each responsible-AI unit, some examples we covered, and the exercises that reinforce the concepts.

\subsection{Designing AI Responsibly (Course 1)}

Access to, adoption of, and use of ML products is inequitably distributed. According to Pew Research, 64\% of Americans believe technology companies create products and services that benefit people who are already advantaged, and 65\% believe these companies fail to anticipate the societal impact of those offerings~\cite{techattitude}. 

To enable more-widespread, safer, and more-secure ML, we must raise awareness of its capabilities. Thanks to the low cost and accessibility of TinyML hardware, our students are diverse, and they will probably have to address different social and cultural factors when designing ML applications. To ensure they all can anticipate the effects of ML products and ensure equitable access, our approach to responsible AI focuses on forming a vision of both the problem to be solved and the people a solution will affect.  

We believe that by taking an active role in responsible ML design, students will be better able to address ethical challenges such as bias, fairness, and security. We therefore cover real-world examples, such as a Winterlight Labs auditory test for Alzheimer’s disease. In this case, research revealed that nonnative English speakers are more likely to be mistakenly flagged as having Alzheimer’s \cite{Fraser2016LinguisticFI}. In a discussion forum, students reflected on what the product designers could have done differently to avoid this failure. Such activities reinforce the importance of considering diverse user perspectives during the design phase, as doing so can inform data-collection decisions that mitigate ML bias. 

In a subsequent forum, learners practice ethical reasoning about the consequences of a KWS model's failure in terms of Type I (false positive) and Type II (false negative) errors. In this case, a false positive would result in audio being recorded, unbeknownst to the user, and sent to the cloud. A false negative means the device failed to activate when the user spoke the wake word. Students must justify their decision to optimize the model for high precision, thereby minimizing false positives, or to optimize for high recall, thereby minimizing false negatives.

For the KWS activity, nearly all participants chose to optimize for high precision to minimize the risk of privacy violations. Interestingly, one provided a justification based on sustainability concerns related to unnecessary data transmission and storage in the cloud. Those who decided to optimize for high recall cited a variety of reasons. One noted that although people claim to value privacy, they tend to prioritize convenience. In contrast, another suggested enacting privacy measures elsewhere to offset the potential harm of optimizing for high recall. Lastly, yet another student prioritized model performance to meet user expectations. That student claimed the burden of preserving privacy should fall on the user, who has the ability to decide whether to purchase the product. There is no right or wrong answer. Our desire is to spur self-reflection and foster constructive discussion among learners from different backgrounds and cultures. 

\subsection{Developing AI Responsibly (Course 2)}

Any developer employing ML must be aware of how data-collection bias and fairness affect application behavior. Our courses use public data sets, including Speech Commands~\cite{warden2018speech}, Mozilla Common Voice~\cite{ardila2020common}, ImageNet~\cite{deng2009imagenet}, and Visual Wake Words~\cite{chowdhery2019visual}, for nearly all of the programming assignments. Most data sets, however, have demographic-representation problems~\cite{pmlr-v81-buolamwini18a}. For example, despite crowdsourcing efforts to increase diversity, the Common Voice data set lacks equal gender representation (only 24\% of English-data-set contributors who revealed their gender are female) \cite{lackoffemalevoicesincommonvoice}. 



Our goal is for students to see how data collection, bias, and fairness intertwine, as well as to equip them to mitigate the problems. because they are working with KWS models, we cover real-life biases relevant to this kind of ML application. For instance, research shows that voice-recognition tools struggle to identify African American Vernacular English, causing popular voice assistants to work less well for black individuals \cite{Koenecke7684}. Similarly, research shows that voice recognition struggles to identify nonnative English speakers and those with speech impairments \cite{10.1145/3379503.3403563}. To acquaint learners with recent work in mitigating bias, we discuss Project Euphonia, a initiative that launched in 2019 with the goal of collecting more data from individuals with speech impairments or heavy accents to fill the gaps in voice data sets \cite{shor2019personalizing}. 

We created a Colab activity that uses Google’s What-If Tool (WIT)~\cite{google_2020_wit}, based on its Responsible AI tool kit~\cite{google_2020}. The WIT is one of the company's many open-source, interactive visualization tools for investigating trained-ML-model behavior with minimal coding. In this exercise, participants practiced ethical reasoning by exploring a real-life data set, identifying sources of bias, and evaluating threshold-optimization strategies for fairness. 
For the WIT activity, students noted how the visual representations fostered a deeper understanding of issues pertaining to fairness. One claimed that the focus on confusion matrices in particular was an effective way to clearly distinguish between the fairness metrics. In general, learners appreciated the opportunity to try out the WIT. 

\subsection{Deploying AI Responsibly (Course 3)}

Even after designing and developing an ML model, deployment raises a new set of ethical challenges. For example, TinyML systems are often touted for preserving privacy. When an embedded system processes data locally (close to the sensor) rather than transmitting it to the cloud, we tend to believe it protects user privacy. But user interaction with the model raises new privacy and security concerns~\cite{prabhu2021privacypreserving}. Moreover, ML interacting with a dynamic real-world environment using sensors raises concerns about model drift\footnote{\emph{Model drift} generally refers to prediction-accuracy degradation owing to environmental changes.} over a product's lifetime.

To the extent that TinyML enables ML everywhere, the privacy, security, and even model-drift risks could be more widespread compared with traditional ML. To familiarize students with these risks, we cover real-life examples, such as doorbells that share data with law enforcement~\cite{doorbelldata} and fitness devices that leak user information~\cite{fitnessleak}.

Our goal is to equip students with strategies to mitigate these risks when deploying trained models in embedded devices. Importantly, the mitigation strategies available to traditional ML systems are sometimes unattractive or infeasible for TinyML. For instance, the resource constraints of an embedded device, such as low power and small memories, complicate implementation of robust security systems and model retraining. Therefore, we acquaint course participants with a wide array of strategies, such as minimizing the transmitted and stored data to preserve user privacy, minimizing hardware design to limit vulnerability to attackers, and running supervised experiments in the real world before releasing ML models.

\begin{figure}[!tbp]
\centering
    \includegraphics[width=\linewidth]{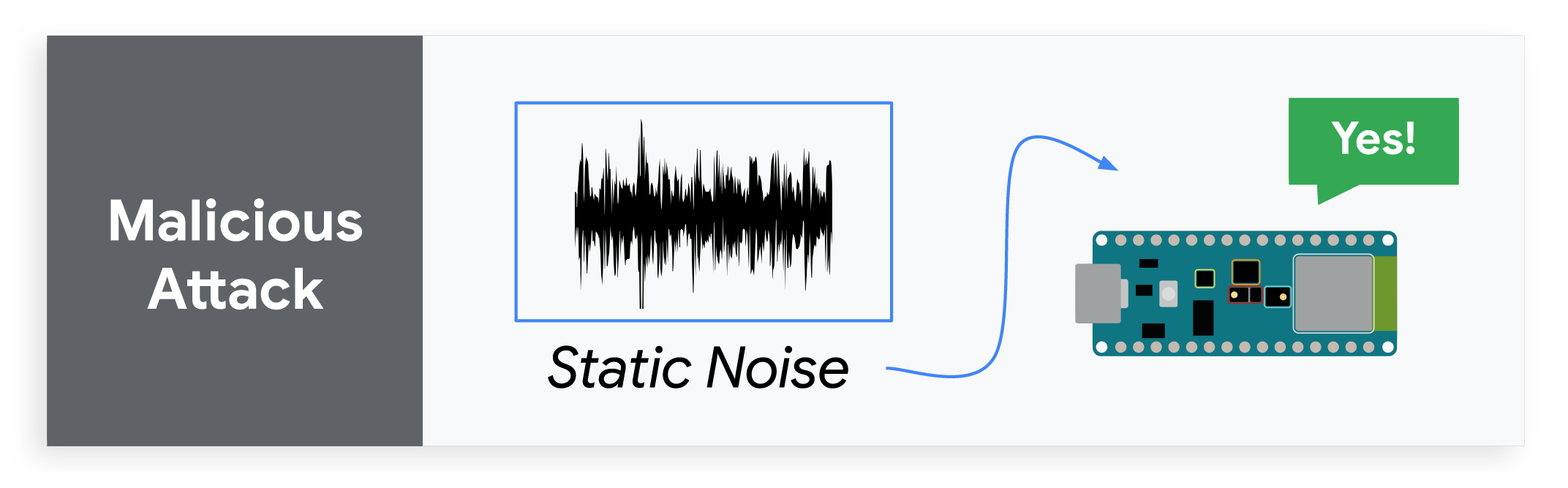}
    \caption{Students attack a pre-trained KWS model with malicious static noise and trigger a spotting of the keyword "yes," showing them the importance of security and privacy.}
    \label{fig:attack}
\end{figure}

Inspired by research showing we can use inaudible ultrasonic noise to trigger or eavesdrop on KWS models~\cite{zhang2017dolphinattack}, we created an exercise that gives students hands-on experience attacking a KWS model. They trigger a false positive---``yes''---with seemingly innocent but adversarial static noise (Figure \ref{fig:attack}), which in a real application would cause the system to constantly record and transmit the audio. This experience builds on our videos and readings and makes the security threat real---a crucial part of any major security-awareness program at large~\cite{reinheimer2020investigation}. At the same time, it is also a cautionary tale of ML's limitations, a lesson all applied-AI engineers should learn. 

To further reinforce this point, a subsequent discussion forum allows course participants to practice ethical reasoning to determine when malicious triggering of a false positive can cause serious harm. Some have noted that this vulnerability would be most likely to cause harm where security is a paramount value, such as using KWS to grant access to a secure space or to initiate a financial transaction. One student drew a connection to the practice of ethical hacking, or penetration testing, and the possibility of developing adversarial data for retraining the model to be more resilient. Interestingly, another noted that since users lack the ability to fix security issues, their only option is to stop using the device. But this choice ultimately depends on whether the company informs customers about the vulnerability. Lastly, one student claimed the adversarial example was more reliable than that student's own voice for triggering a ``yes.'' The course staff then responded, prompting a discussion of data-set bias and the likelihood that American male accents are overrepresented in the data.

\subsection{Scaling AI Responsibly (Course 4)}

Many ethical implications require consideration when applying technology. Even minor biases, which can be difficult to detect in the proof of concept, can have a major impact when appearing in thousands or millions of devices.



This problem highlights the need to treat responsible ML as an iterative process. Rather than introduce entirely new ethical considerations, we revisit and expand on previous ones. For instance, to guide students in cleaning up a test data set before they conduct benchmarks, we revisit the ethical issues of data collection and bias. Similarly, we revisit privacy in depth once participants become acquainted with federated learning.

We are incorporating an active learning exercise using Google's Model Card Toolkit (MCT)~\cite{Mitchell_2019}. Model cards are a reporting mechanism that can increase model transparency and facilitate the exchange of information between model creators, users, and others. This exercise requires that students practice using the model-card framework to document information relating to the model's development, performance, and ethical considerations. 

We additionally discuss the environmental impact of large-scale TinyML networks, as the production and maintenance of billions of MCUs can have lead to substantial carbon emissions. Beyond the ethical pitfalls of scaling TinyML, we cover the potential positive social impact this technology can have in domains such as environmental sustainability, public health, and AI equity.
\section{Access via MOOCs} \label{sec:atscale}

In this section we describe how we leveraged technology to make the TinyML specialization broadly accessible and highlight important considerations made to ensure we supported our remote learners. 

\subsection{Massive Open Online Course}

Our goal was to reach a global audience. We therefore chose to employ massive-open-online-course (MOOC) platforms. Examples such as edX and Coursera are ideal for making the content globally accessible; students need not travel to a different country to learn. These platforms host a wide variety of university-level courses and are generally cheaper than equivalent academic and professional training thanks to the economics of scale~\cite{belleflamme2014}. We deployed the TinyML specialization on edX through HarvardX. 

Participants can audit the course for free or pay to earn a professional certificate. Since they can ``upgrade'' to a professional certificate at any point during the course, both students and professionals can try before they buy, encouraging more to enroll. Although the professional certification includes summary tests that are absent from the audit version, we designed the curriculum so individuals who are just auditing learn the same crucial principles. Thus, all can attend the entire class, developing their skills for free. At the time of this writing, the number of auditors far outweighs the number of paying students by more than order of magnitude. 

The course is asynchronous and self-paced rather than instructor led. Students progress through the material at whatever speed they find comfortable. But unlike in-person courses, interaction between students and staff is minimal, forcing staff to develop high-quality, self-explanatory, and self-sufficient materials that rely heavily on media (which we describe in greater detail in Section~\ref{sec:media}).



Unlike most MOOCs, Course 3 employs the TinyML kit (Section~\ref{sec:kit}) for hands-on learning. To maximize hardware accessibility, we worked with Arduino to make a custom all-in-one kit globally available for purchase through either that company's website~\cite{arduino} or one of its many distributors. We also provided a detailed bill of materials for students who wish to buy individual components instead. The main benefit of this approach is that it improves the efficiency for the host institutions (Harvard and Google) by reducing the burden on them for managing inventory and shipping logistics (taxes, international shipping rates, etc.).









\subsection{Accelerated Remote Media Production}
\label{sec:media}

The typical development timeline for a series of online courses, such as the ones we described in Section~\ref{sec:courses}, is about two years---far too long to keep up with changing ML technology. Applied ML, especially in the context of TinyML, remains a nascent yet quickly developing field. Therefore, media production for the online curriculum must be rapid to ensure the material is timely and relevant and to ensure broad access.

\setlength{\fboxsep}{0pt}
\begin{figure*}[t!]
     \centering
     \begin{subfigure}[b]{0.24\textwidth}
         \centering
         \fbox{\includegraphics[width=\textwidth]{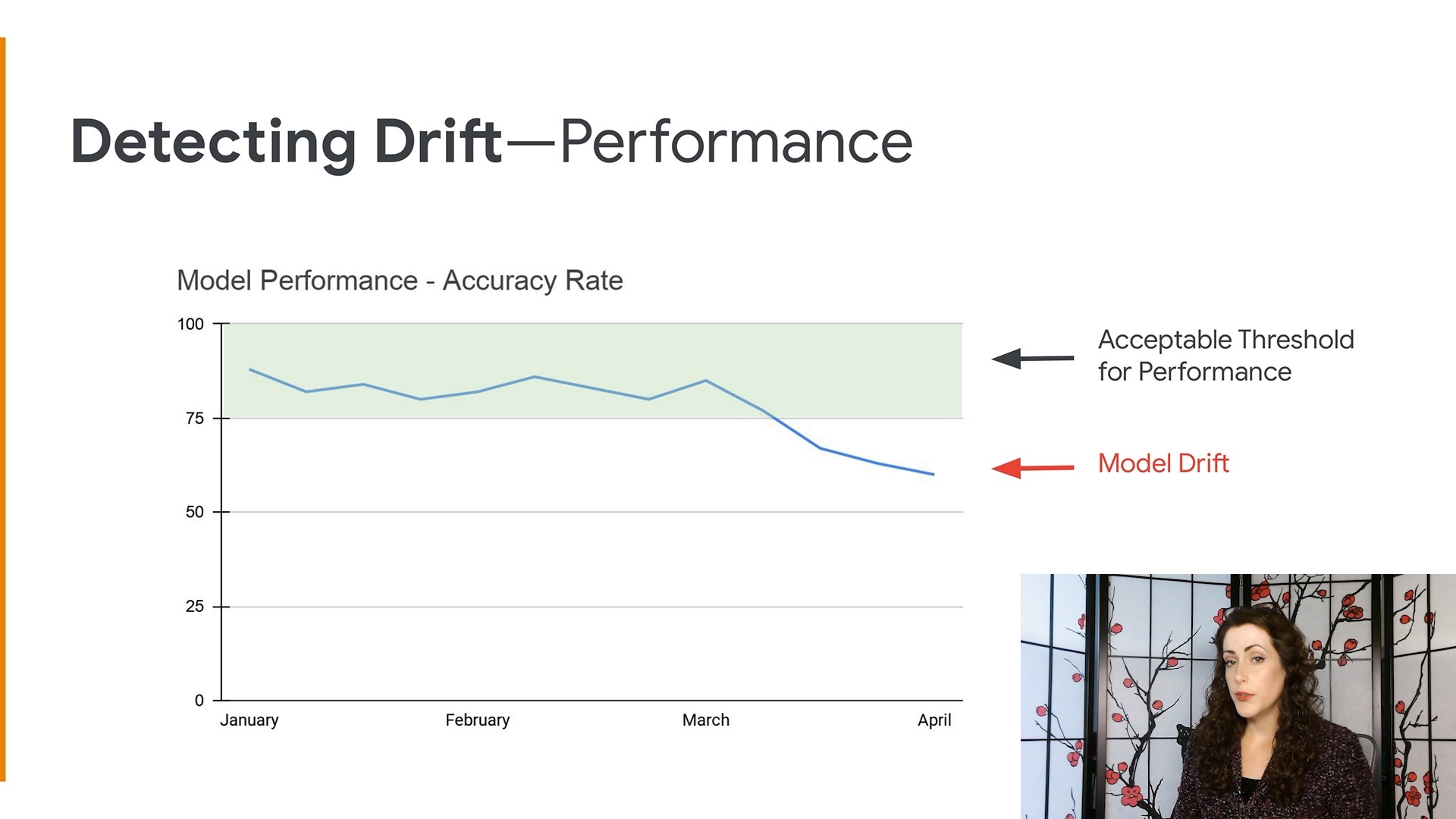}}
         \caption{Picture-in-picture}
         \label{fig:pic-in-pic}
     \end{subfigure}
     \hfill
     \begin{subfigure}[b]{0.24\textwidth}
         \centering
         \fbox{\includegraphics[width=\textwidth]{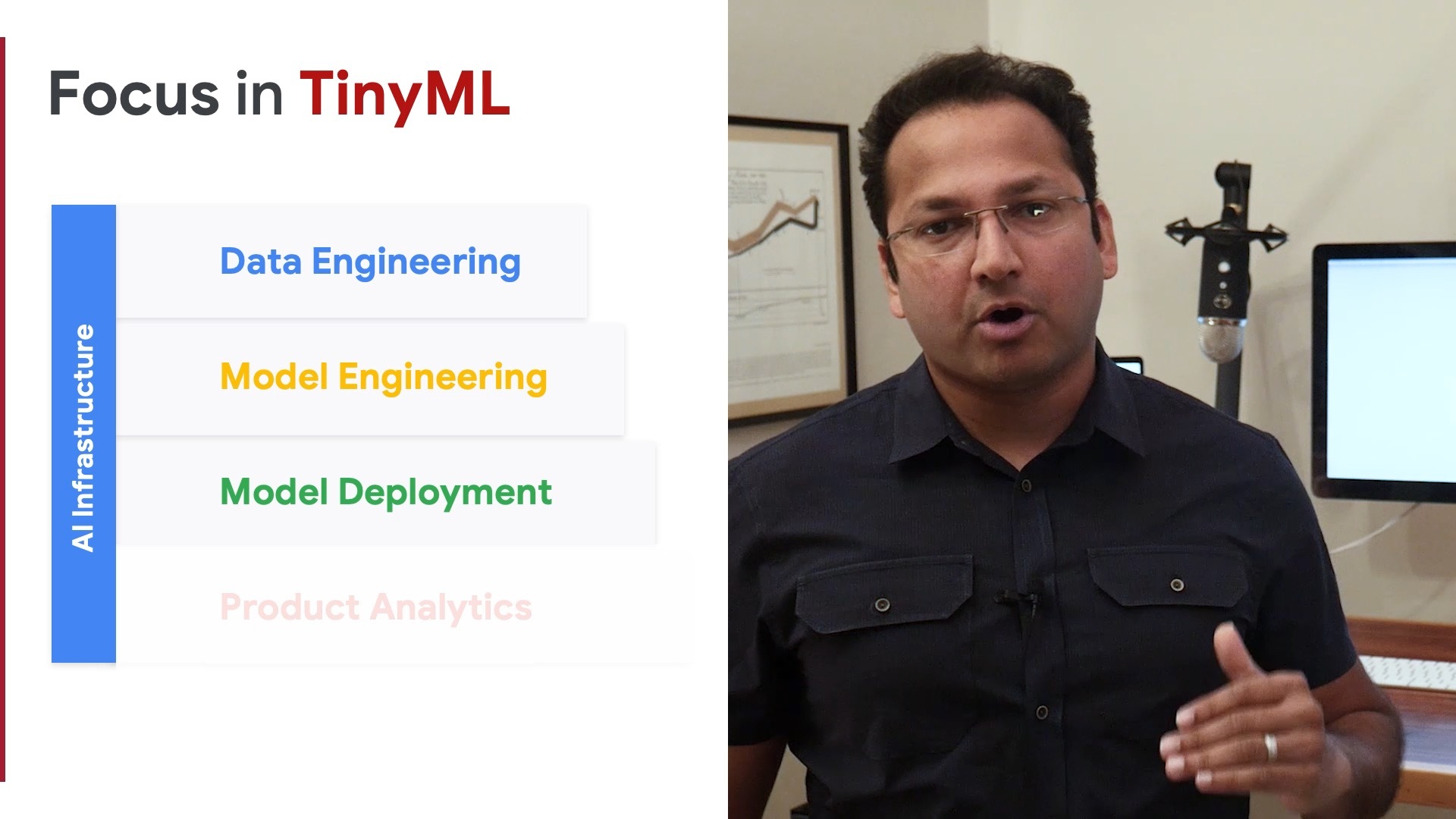}}
         \caption{Split screen}
         \label{fig:split-screen}
     \end{subfigure}
     \hfill
     \begin{subfigure}[b]{0.24\textwidth}
         \centering
         \fbox{\includegraphics[width=\textwidth]{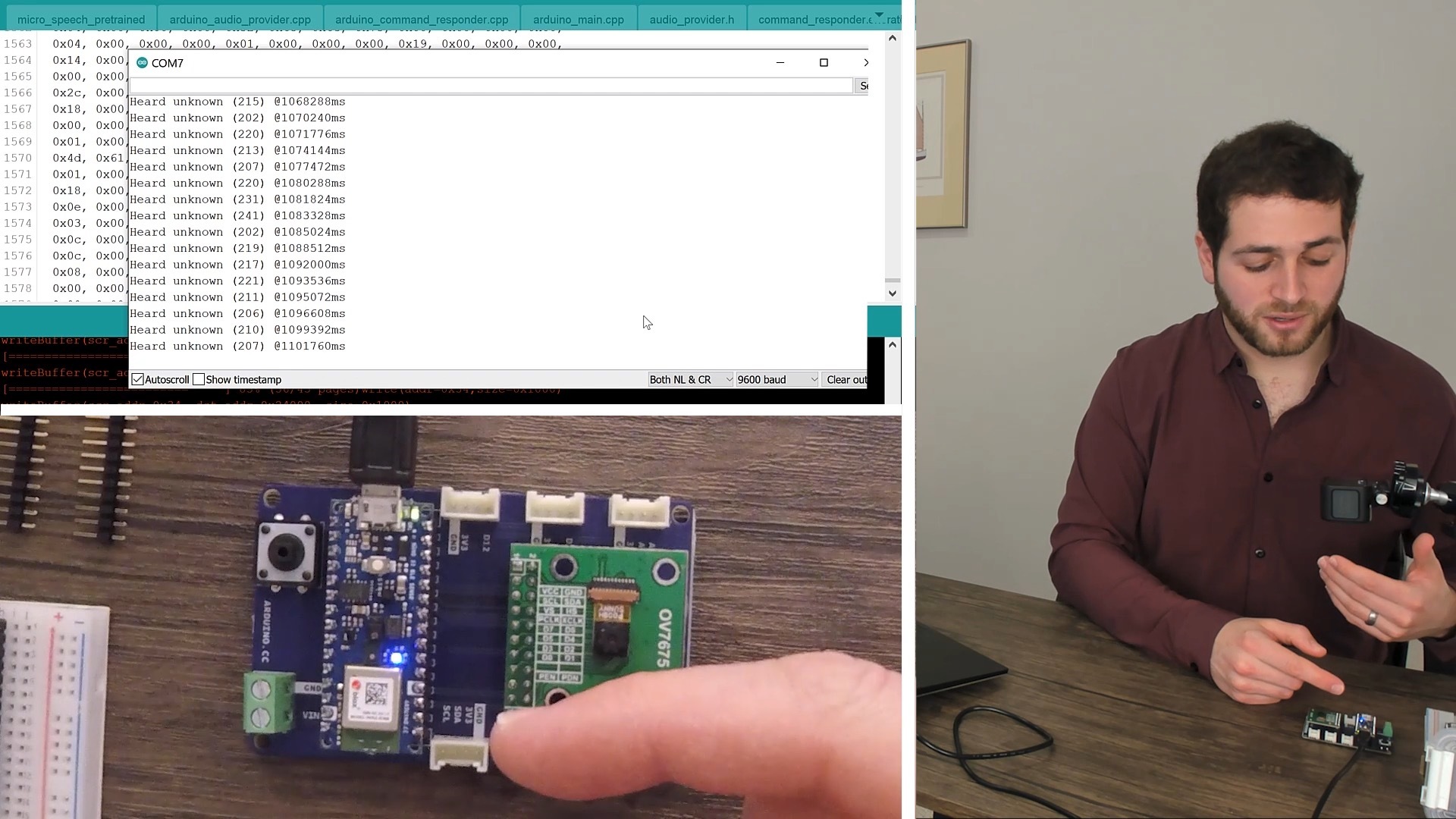}}
         \caption{Tri-frame for lab demos}
         \label{fig:lab}
     \end{subfigure}
     \hfill
     \begin{subfigure}[b]{0.24\textwidth}
         \centering
         \fbox{\includegraphics[width=\textwidth]{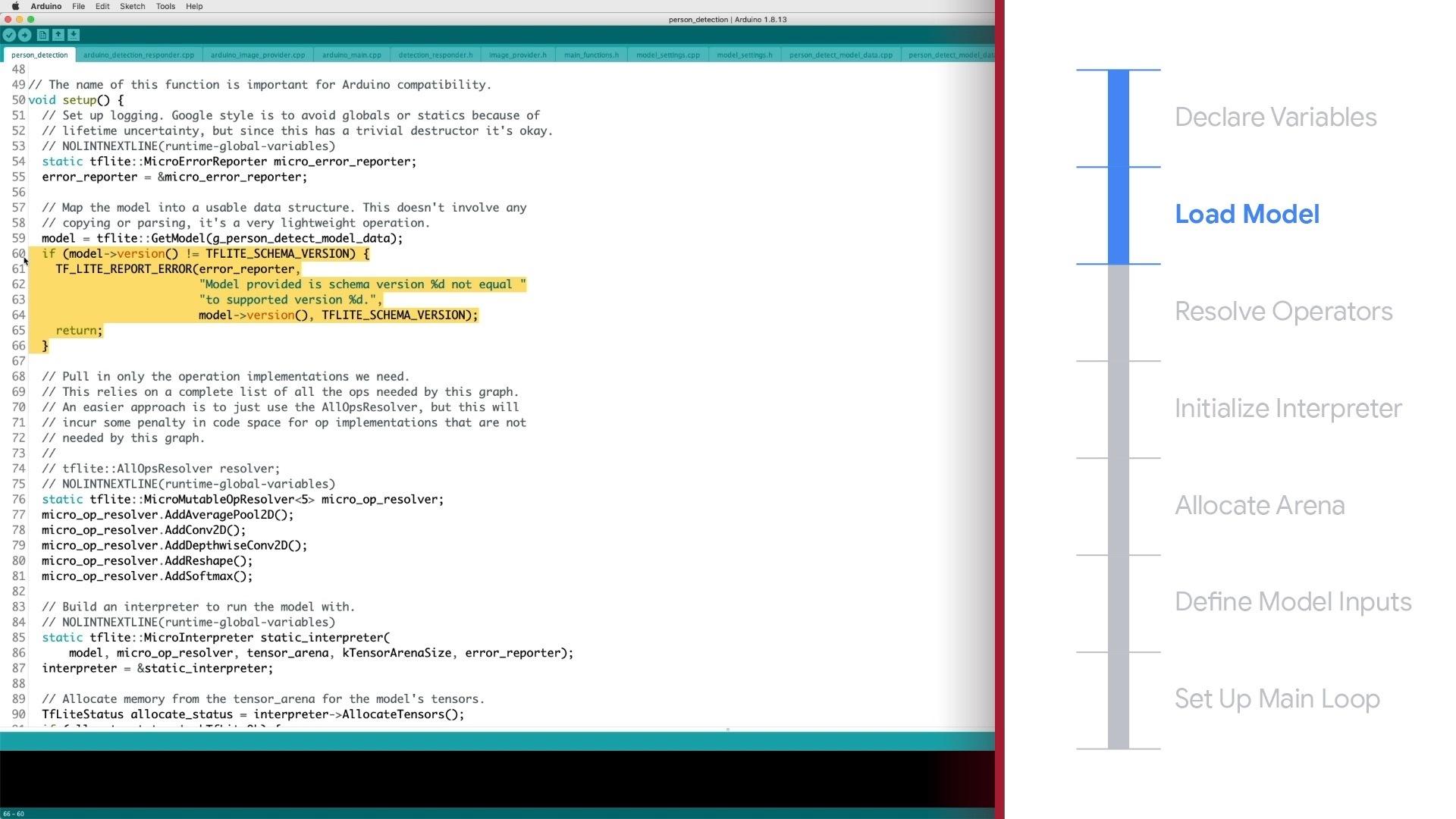}}
         \caption{Screencast}
         \label{fig:screencast}
     \end{subfigure}
    \caption{We used various video-production strategies throughout the course to maximize learning efficiency. (a) Picture-in-picture places a video clip in a small frame on top of another frame, playing them simultaneously. It enhances the perception of instructor presence while showing the student relevant content. (b) Split-screen, a slight variation, also improves the perception of instructor presence. (c) Tri-frames are useful for lab demos to enable ``hands-on'' instruction from teaching assistants. (d) Screencasts coupled with associated lecture material reinforce concepts with code.}
    \label{fig:media-types}
\end{figure*}

We compressed the media-production time greatly, achieving an average development cadence of 6--8 weeks per course. To maintain this cadence, we created a custom remote-media-production workflow. We produced the TinyML course under the specter of Covid-19, but regardless of the safety limitations, a remote production strategy would still have been the only way to achieve these quick results. Remote production methods offer flexibility and allow an international crew to make contributions, meaning the process continues around the clock. Regardless, no matter when and how it is done, creating a flexible workflow requires a principled content-design approach, and advanced technology is necessary for rapid progress. The following breakdown can serve as a roadmap for others attempting to follow a similar approach.

\textbf{Production design.} To expand access such that our effort meets the needs of a global audience (Section~\ref{sec:pedagodgy}), we built our media-production strategy around five critical ingredients: compelling instructional narrative, best media practices for online learning, a diverse and skilled production team, prioritized use of production equipment, and willingness to innovate. 

A \emph{compelling instructional narrative} that whets student interest is critical, as all great media experiences unite around a good story. 
TinyML offers a sound narrative because it provides an accessible, hands-on introduction to ML (Section~\ref{sec:whytiny}). We aimed to communicate with a global audience and provide the practical knowledge for building complete, relevant TinyML applications and tools.

Effectively communicating that narrative requires \emph{best media practices for online learning}. Decisions made in postproduction often hold more weight than any others. For example, the decision of when to show the instructor, slides, or both in a picture-in-picture cut versus when to display graphics or other visual/auditory information can affect the viewer's cognitive load and overall learning~\cite{chen2015effects,mayer2020five}. When in doubt, Mayer's ``12 principles of multimedia learning''~\cite{mayer2005cambridge} is an excellent place to discover such general practices for enhancing the student's experience.

From the start, we determined the primary media types we would produce to hold students' attention and maintain their cognitive load balance~\cite{chen2015effects}. We chose picture-in-picture and split-screen formats, allowing us to show the instructor or other imagery in full-screen mode to focus on the most important aspects of the presentation (Figures~\ref{fig:media-types}). We emphasized instructor screen time, however, to improve student learning~\cite{wang2020does}.

A \emph{geographically distributed and responsive team} is necessary to quickly produce highly sophisticated content, especially for an emerging technical field. Our media team included a producer and director to establish a creative vision and ensure media delivery, a senior editor to assemble and craft the videos, a motion-graphics designer to provide custom graphical elements for our brand, and a production assistant to wrangle data, review content, and integrate the final videos into the platform. This team was relatively lean. One additional advantage was that contributors were scattered across 12 time zones (San Francisco to Boston to London to Mumbai), meaning at all times someone was awake and working on the project. 

A crucial ingredient to quickly producing content is \emph{prioritized use of production equipment}.
The remote nature of the production and the Covid-19 precautions only heightened this need.
For example, webcams and audio supplies were sold out or on back order because people were setting up home offices so they could continue to work. Fortunately, we were able to make acceptable compromises and buy equipment in a way that ensured the greatest impact. We prioritized production-equipment purchases as follows: 1) audio, because it is more important to retaining viewer attention than video~\cite{AudioVideo2014}; 2) lighting, as it can improve even a nonideal camera to draw the student's eye; and 3) video, which we mention last because it is the most expensive in a context where higher production value does not necessarily imply a better learning experience~\cite{guo2014video}.

The final ingredient was \emph{willingness to innovate}. Course~3 (Deploying TinyML) involves hands-on learning. Typically, in-person teaching assistants (TAs) demonstrate labs to show students the goals and scope and to preemptively troubleshoot common errors. Doing the same online is extremely difficult. We developed a three-way split-screen medium (Figure~\ref{fig:lab}) that displays the device assembly, device testing, and TinyML lab exercises. We assembled a new film location to (remotely) support the teaching staff with the lab exercises, adding an overhead camera and additional lighting. Furthermore, we enhanced our visuals for the three-way split screen with a custom motion-graphics layer. This setup reached completion and underwent rapid testing without disturbing the production timeline. From start to finish, Course 3 took only eight weeks despite involving five hours of produced-video time, which includes short lectures, screencasts, and lab videos.

\textbf{Technology.} Without globally accessible technology and services, remote media production at the level and pace we achieved would have been impossible. Cloud storage was the backbone of our strategy. It allowed contributors to ingest and manage footage globally. It was also the heart of our production workflow, giving us the ability to sync media project files instantly. Videotelephony services such as Zoom and Google Meet aided in assessing home-studio setups in addition to serving as a virtual rehearsal stage and writers' room. Amazon supplied 90\% of our equipment. Frame.io streamlined our video-quality review and revision~\cite{frameio}. 


\textbf{Copyright.} Although on-camera presence was a major focus of remote production, video lectures are just one part of the students’ activities. At the same time, a multidisciplinary team of content experts, graphic designers, and web developers at HarvardX rapidly designed and formatted readings and coding exercises. A major challenge in quickly producing course materials was ensuring each illustration, photo, and code library met strict licensing requirements to avoid copyright infringement. Given our project's more than two thousand graphics and tight timelines, we trained all content developers on proper sourcing for course materials. In-house custom graphics---necessary for a nascent field---predominated, and copyright specialists at HarvardX evaluated each piece as it arrived to cite all external creators.

\subsection{Building Community}

A common and well-known pitfall of MOOC platforms is the difficulty of developing community and fostering peer learning among a geographically distributed population. Students often struggle to discuss and collaborate after completing the course and even during the course. We therefore developed the TinyMLx community, which welcomes everyone beyond the edX platform.

First, we created a Discourse forum (\href{discuss.tinymlx.org}{discuss.TinyMLx.org}) to provide both a communication platform for students and a home for future initiatives. It has been successful, garnering over 3,500 user visits and over 58,500 page views in its first five months. We also conducted two live Q\&A sessions for the TinyML community. For each session, between 100 and 200 learners joined live from around the world, and many more have since watched the recording. We received dozens of questions leading up to the events and dozens more during, with topics including how to best teach TinyML material, how to improve diversity in TinyML, and many others in between. Participants enjoyed the events, e.g., 90\% of respondents to our first post-event poll said they would like to attend another. Finally, based on learner feedback we recently created a Discord chat to further enable easy collaboration, communication, and community building.

To challenge our Course 3 students, who went on to deploy TinyML models on their microcontrollers, we developed an optional ``capstone-project'' competition. We believe this competition reinforces the value and usefulness of the technical skills that students are gaining. A prize will go to the individual (or individuals) whose project demonstrates technical mastery, is most creative in its implementation, and has the most potential to improve society. This initiative has already spawned collaborative-learning groups. 

To increase the impact of these projects and further reinforce the real-world applicability of the knowledge students gain through this course series, we are working with the Arribada Initiative~\cite{arribada} to create larger advised projects. This partnership will allow students to contribute their newly acquired TinyML skills to real-world conservation efforts, such as human/elephant conflict mitigation and sea-turtle monitoring, while receiving advice and support from both industry professionals and course staff. Finally, we are asking the community to continuously improve the course, since it is as much theirs as ours. As a result, we've seen many forum posts and GitHub pull requests offering typo corrections, bug fixes, and even content-improvement suggestions.

\begin{figure}[t!]
    \centering
    \fbox{\includegraphics[width=\columnwidth]{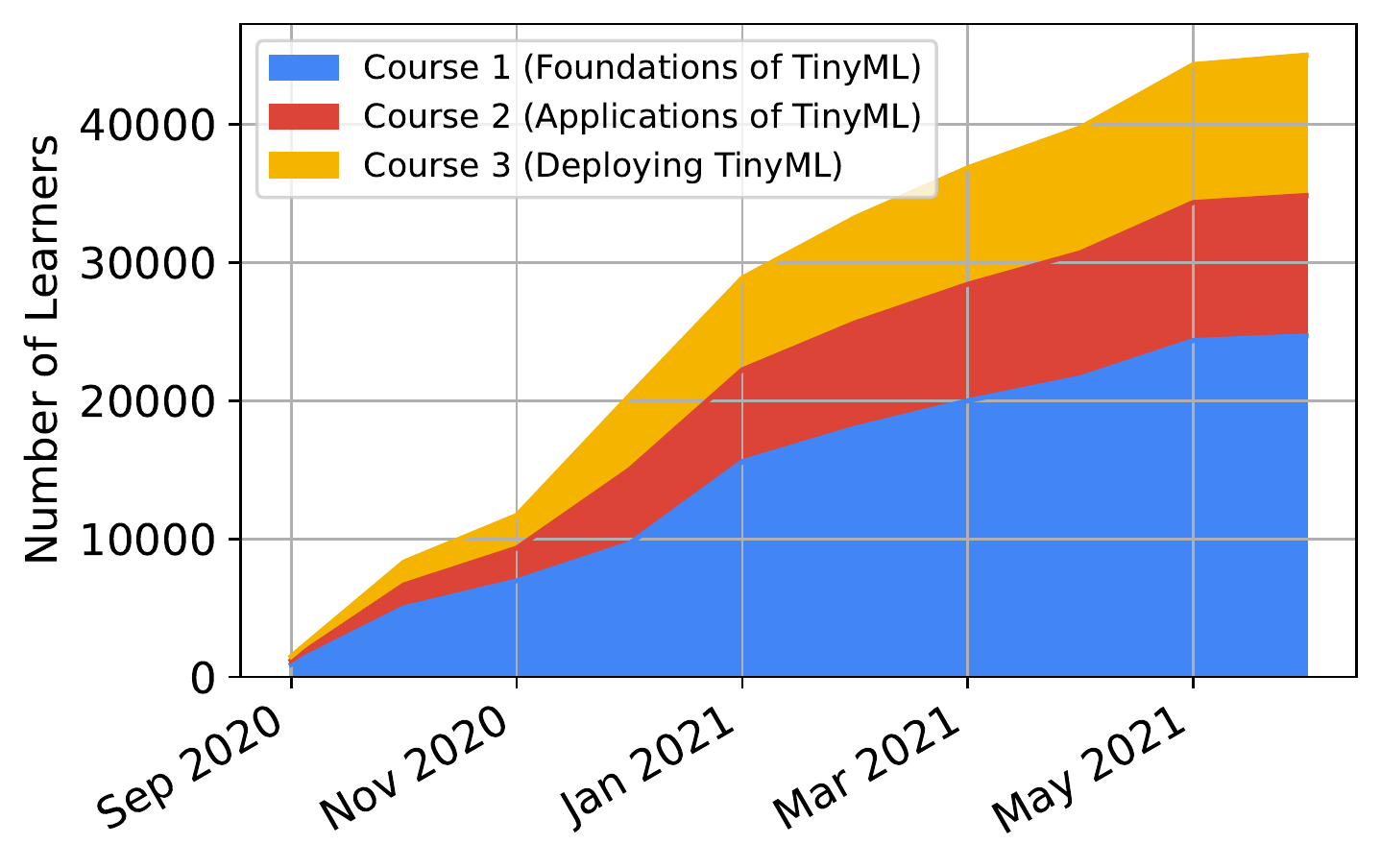}}
    \caption{Course-enrollment metrics for Courses 1 through 3. Over 43,000 students are currently enrolled across all three courses. Course 4 has yet to open for enrollment. We expect another enrollment spike with ``Scaling TinyML.''}
    \label{fig:enrollment}
\end{figure}

\section{Broader Impact} \label{sec:evaluation}

Our goal is to expand global access to applied ML through the lens of TinyML. In this section, we assess our work's initial impact by presenting data from edX Insights, a service that provides course statistics to instructors and staff. It is merely an initial impact assessment, as the first cohort of participants have just begun graduating from the core TinyML series (Courses 1-3), and Course 4 (optional) remains in development. As such, our early analysis considers enrollment in the first three courses by geography, background, age, and gender. 


    
    
    
    

\subsection{Course Enrollment}
At the time of this writing, the total course enrollment stands at 43,000. Figure~\ref{fig:enrollment} shows the daily enrollment data, starting from the opening date. We announced Courses 1, 2, and 3 together in early October 2020 and launched them on October 27, 2020; December 16, 2020; and March 2, 2021, respectively. Students could enroll in any or all courses at the same time but could only start after each course's launch date. 

TinyML is a young field, so the first useful metric is interest in the topic (i.e., acquiring applied-ML skills via TinyML). Figure~\ref{fig:enrollment} shows the strong and steady increase over time. On average, $\sim$1,000 new students enroll in at least one course each week. Interest in Courses 2 and 3 continues to grow---a phenomenon we attribute to participants promoting them through social media such as LinkedIn, Twitter, and Facebook as they earn their course-completion certificates. The sharp increases around the first week of October, third week of December, and third week of February align with course-announcement dates or major social-media activity. For instance, on January 24, Mashable handpicked ``Fundamentals of TinyML'' as one of the 10 best free Harvard courses to learn something new~\cite{turner_2021}. TinyML ranked at the top of the STEM-courses listed.

\begin{figure}[t!]
    \centering
    \fbox{\includegraphics[width=\columnwidth]{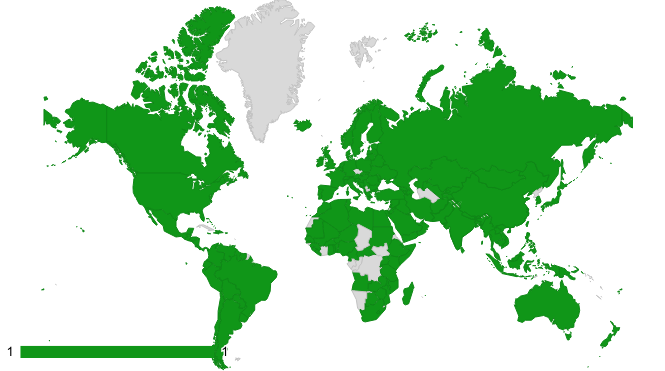}}
    \caption{Global access to TinyML courses. At the time of this writing, people from more than 171 of the 193 United Nations member states have participated in TinyML.}
    \label{fig:global}
\end{figure}

Figure~\ref{fig:global} shows that TinyML students come from more than 171 countries. Because edX reaches a wide audience, our learners come from nearly all continents. Today, the top 10 countries by participant activity are the US, India, Turkey, the UK, Canada, Pakistan, Germany, Brazil, Australia, and Indonesia.

\subsection{Completion Rates}

People take online courses for a wide variety of reasons. Some are curious about the topic and want to get their feet wet; they may audit a course but not complete it. Others would like to master the program and earning a certificate of completion, assuming they can afford it. Therefore, enrollment numbers alone are insufficient.

We assessed how many verified enrollees complete the course. We have access only to the percentage who have earned a passing grade among those officially enrolled in the courses (i.e., the paid-certificate program). This number is constantly changing. At the time of writing, the completion rates are 59\%, 55\%, and 44\% for Courses 1, 2, and 3, respectively. We believe Course 3's number is slightly lower because it is more challenging than Courses 1 and 2, which do not have a hands-on component. The average completion rate for most MOOCs is somewhere between 5\% and 15\%~\cite{hollands2019benefits}, so the TinyML courses appear to be faring well. Although these results are preliminary (we need more data to make better quantitative comparisons), they shed a positive light on our design approach. 

\subsection{Learner Demographics}

We conducted a demographic analysis of students' age, educational background, and gender. They volunteer this information to edX, so it covers only a fraction of the numbers in Figure~\ref{fig:enrollment}. Nonetheless, the data is extensive enough that we can draw general conclusions. At the start of each course, a forum post asks students to introduce themselves and summarize what they hope to get out of the edX series. We derived additional qualitative analysis from these responses. So far we have a good distribution across age groups and educational backgrounds. Our gender diversity is lacking, however, but we are working to address it (Section~\ref{sec:future}).

\begin{figure}[t!]
    \centering
    \fbox{\includegraphics[width=0.9\columnwidth]{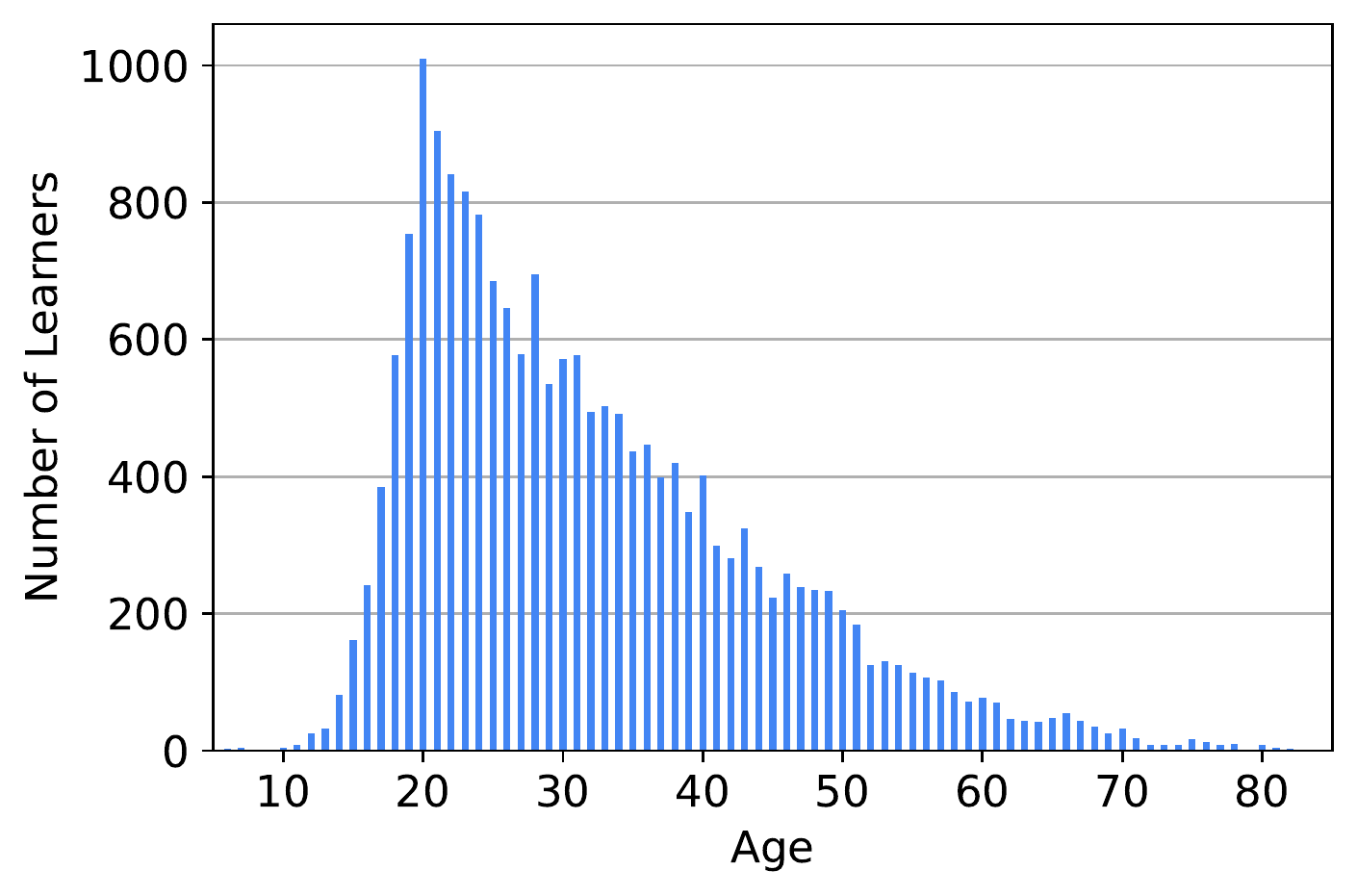}}
    \caption{Age demographics across all courses based on voluntarily provide information. We have learners who are still in high school to learners who are retired and learning TinyML to understand its impact on our global society.}
    \label{fig:age}
\end{figure}

\textbf{Age.} Figure~\ref{fig:age} shows the age distribution for all three courses combined. The median is 30. Some participants are high-school students as young as 15 and wish to pursue an ML career. Others are over 60 and wish to understand the latest technological innovations as well as their societal implications. This age diversity was one of our objectives (Section~\ref{sec:pedagodgy}).

\begin{figure}[t!]
    \centering
    \fbox{\includegraphics[width=0.9\columnwidth]{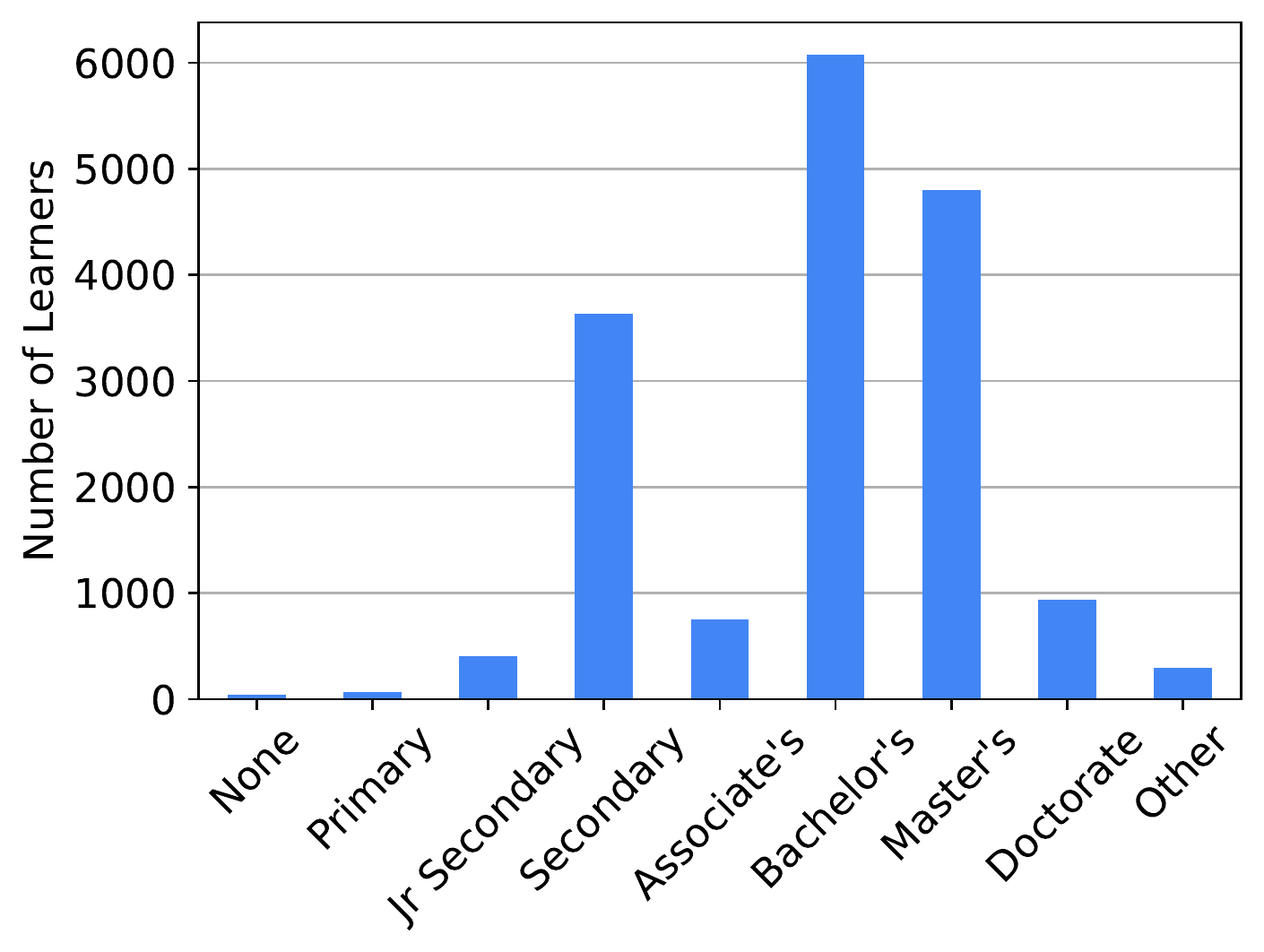}}
    \caption{Education demographics based on voluntarily provide information. Many of our learners indicated an interest in TinyML to understand applied ML technologies to either pivot or grow further in their current positions.}
    \label{fig:education}
\end{figure}

\textbf{Education.} Figure~\ref{fig:education} shows that nearly all our learners have either just a secondary (high-school) diploma or a bachelor's/master's degree. A few others have doctorate degrees. Judging from the forum discussions, we gather that individuals with a bachelor's or master's degree are trying to advance or shift their careers by adding an ML focus. Most participants with a doctorate want to apply (tiny) ML in their research. Many students expressed enthusiasm about enrolling in a career-advancing course backed by both Harvard and Google. This variety of educational backgrounds and career focuses also meets our expectations and objectives and further emphasizes the importance of our academia/industry partnership (Section~\ref{sec:pedagodgy}).

\begin{figure}[t!]
    \centering
    \fbox{\includegraphics[width=0.9\columnwidth]{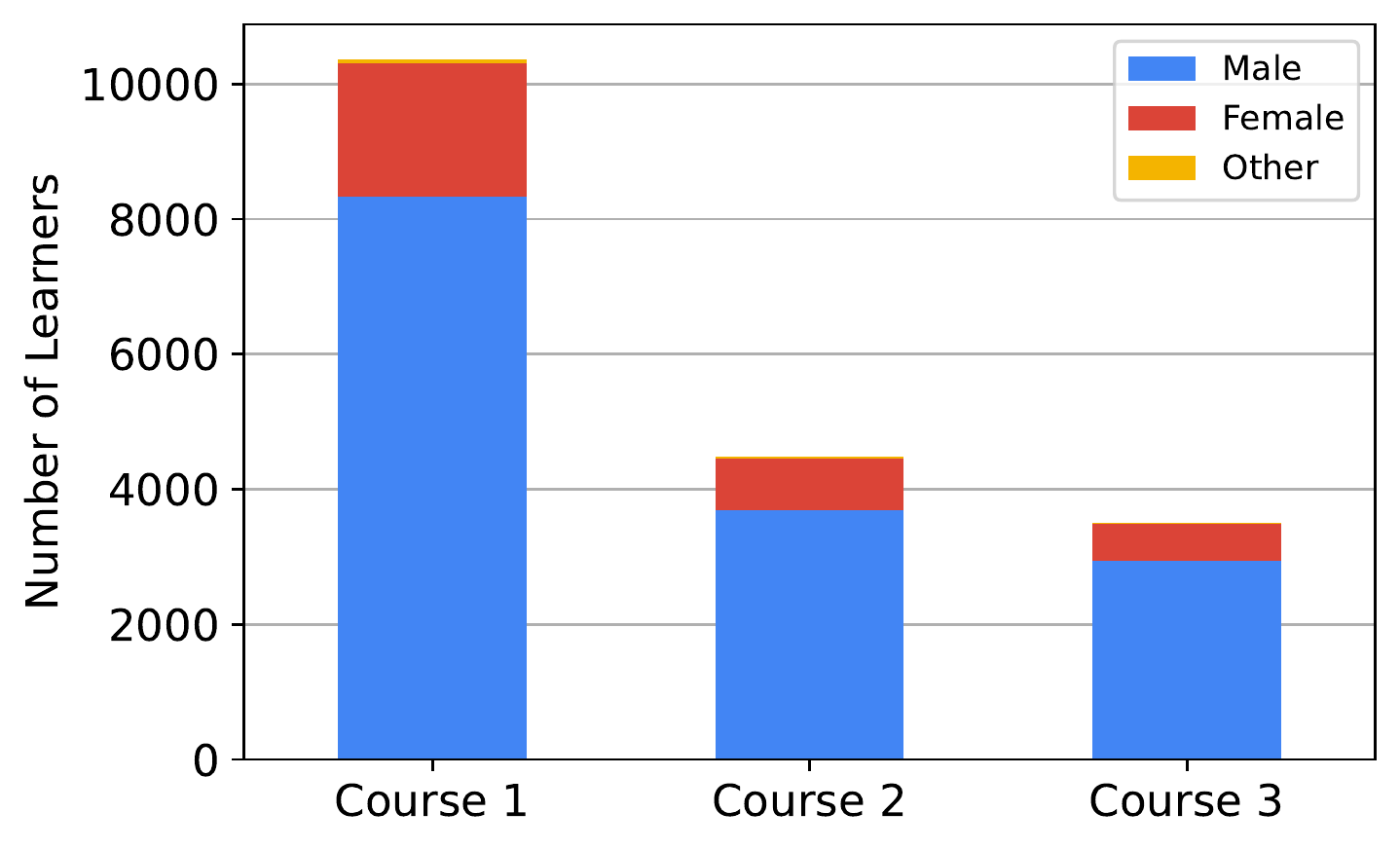}}
    \caption{Gender demographics based on voluntarily provide information. Collectively, we are working on engaging a more diverse population of learners with the aid of working groups that are part of the tinyML Foundation.}
    \label{fig:gender}
\end{figure}

\textbf{Gender.} Figure~\ref{fig:gender} depicts the gender diversity across all three courses. It weighs heavily toward men; on average, across all three courses, 20\% of our learners are women. We are working to change that ratio through our open education initiative (Section~\ref{sec:future}). More specifically, we are putting together a TinyML4Everyone working group to encourage more women to learn about TinyML.



\section{Future Directions} \label{sec:future}

TinyML can dramatically transform applied-ML education and development at many levels, far beyond what we achieved with the edX specialization. To this end, we launched the Tiny Machine Learning Open Education Initiative (TinyMLx)~\cite{tinymlxopen} to sponsor a wide variety of initiatives, such as TinyML4D (for applied-ML education via TinyML for developing countries), TinyML4STEM (for nurturing creative research in science, technology, engineering, and math), TinyML4Everyone (for building a shared identity and breaking stereotypes), and TinyML4x (for your favorite topic \textit{x}).

We are currently running the TinyML4D and TinyML4Everyone working groups that are looking for ways to broaden TinyML participation, access, and belonging. One way is to provide TinyML materials in a student's native language. For instance, we already have two projects for developing course content and instructional materials in Spanish and Portuguese~\cite{rovai21}. Additionally, the makers of TinyML on edX, along with students and faculty of Navajo Technical University in New Mexico, plan to conduct a workshop in June 2021 that teaches Navajo students the basics of hardware programming and how to employ ML for their communities by creating voice-activated applications trained on the Navajo language.

TinyML can be instrumental for inspiring youth, as it offers a superb introduction to programming and ML for K--12 students. Deployment of TinyML applications on physical embedded devices intrigues students by allowing them to interact with actual technologies, not just on-screen representations. 
Our first step in this direction was to publicly release all course materials on our GitHub: \href{https://github.com/tinyMLx/courseware}{\texttt{https://github.com/tinyMLx/courseware}}. We are working with STEM teachers worldwide to help us refine our tools for the classroom. Our aim is to develop ready-to-go project-based lessons and accompanying lesson plans to further increase ML access by reaching younger children. One possible project is to enable the use of visual programming abstractions (e.g., \href{https://maker.makecode.com/#editor}{Microsoft MakeCode editor} for the Arduino Nano BLE Sense 33) so people of all ages can apply ML without learning a programming language.

In addition, we are working with various organizations to assist teachers in learning applied ML. The 2021 Backyard Brains AI Fellowship~\cite{backyard_brains_2021}, for example, is an early opportunity for teachers to help design TinyML projects for classrooms.


\section{Limitations}
\label{sec:limits}





We believe TinyML is an effective means to widen access to applied ML. Indeed, it is one way but not the only way. To provide a more balanced viewpoint, we describe some limitations of our approach and suggest alternative methods that may be more suitable. 

\textbf{Hardware cost.} TinyML requires the purchase of embedded hardware to acquire the full-stack ML-development experience. The TinyML kit we developed costs US\$49.99. In some developing countries, this exceeds the average income in a week, in some rare cases, even a month. Although this price is considered reasonable in some countries, it may still be too high in others. We have found that the cost of shipping to distant parts of the world depends heavily on the presence of nearby distribution centers that carry the device. If none exist, the kit's cost, including shipping, can sometimes double the original kit price.

Ideally, TinyML would require no physical hardware, making the hardware cost zero. We are experimenting with open-source emulation platforms such as Renode.io from Antmicro~\cite{antmicro}. Renode is an open-source framework that allows users to build and test embedded (ML) software without physical embedded hardware. It will enable developers to run their original code, which would have run on the hardware, unmodified in an emulated environment. Although this approach eliminates the hardware cost, students miss the opportunity and excitement of interacting with a device.

\textbf{Device accessibility.} Globally, the number of embedded devices far exceeds the number of cloud and mobile devices (as Figure~\ref{fig:tinymlcount} shows). But individuals must procure the necessary embedded hardware, such as the TinyML kit that we have developed with Arduino, to learn. By comparison, devices such as laptop and desktop computers connected to the web benefit from easier access. Students can use a regular computer to gain access to the online course materials. Even if they lack immediate access to computers in their homes, they can access the online resources from Internet cafés that provide web access for a nominal fee. A crucial shortcoming of this approach, however, is that learners will have difficulty experiencing the complete ML workflow (Figure~\ref{fig:workflow}), since they will be unable to deploy in a device the models they train in the cloud.

Smartphones may be a suitable compromise. They are highly accessible, even though they can be an order of magnitude more costly than the TinyML kit. Nevertheless, they enable students to experience the complete TinyML design, development, deployment, and management workflow. Also, an average smartphone has more than 10 sensors---many more than the Arduino Nano 33 BLE Sense we use in Course 3, enabling additional applications. Learners can hold the smartphone in their hands, much like the TinyML device. 
That said, 
conveying the significance of ML's future being tiny and bright (Section~\ref{sec:whytiny}) is more challenging (though not impossible) because mobile devices have far more resources (compute power, memory, bandwidth, etc.) than TinyML devices (Table~\ref{tab:mcu_scale}). Students may therefore miss the fundamental issue of embedded-resource constraints. But if the goal is ultimately to expand access to applied ML, mobile devices may be a fair compromise.

\textbf{Programming background.} Building ML models for mobile devices (using TensorFlow Lite~\cite{tflite}) or the web (using TensorFlow.js~\cite{smilkov2019ensorflowjs}) is possible using high-level programming languages such as Python and JavaScript, respectively. These languages are easy to learn and far more accessible to beginners than C/C++, which is necessary to program embedded hardware (similar to Course 3). So although TinyML creates an opportunity to showcase the full-stack ML experience using embedded hardware, and we leverage the Arduino IDE and heavily scaffolded code with video walkthroughs to minimize the lift to C/C++, it may also narrow access in some regards. The additional necessary programming skills and associated education can be a roadblock.

In the future, we believe that end-to-end developer platforms such as Edge Impulse~\cite{EdgeImpu51:online,Introduc46:online} that lower the entry barrier into TinyML will likely become mainstream and an essential part of the future developer ecosystem. Not every embedded ML engineer must know and understand all of the inner workings of TensorFlow Lite Micro or how an ML compiler works or how to extract the best performance from a highly customized ML hardware accelerator etc. Instead, learners need the right level of abstraction that allows them to focus on what matters most. Platforms such as Edge Impulse make it easy for learners, software developers, engineers and other domain experts to solve real-world problems using ML on the edge and TinyML devices without advanced degrees in ML or embedded systems. We therefore expose learners to the end-to-end MLOps platforms in Course 4, but note that more focus on such platforms in future courses could enable even more accessibility.

In summary, there are many paths to broaden applied-ML access. The correct approach--or, better, the most suitable approach--depends on the situation. We, therefore, hope this discussion clarifies the pros and cons of approaching applied ML through TinyML.
\section{Conclusion}
\label{sec:conclusion}

Expanding access to high-quality educational content, especially for machine learning, is important to ensuring that expertise diffuses beyond just a few prominent organizations. But doing so in a way that is both accessible and affordable to many different people is a difficult task. The four-part TinyML edX series we present here aims to tackle these challenges by providing application-driven content that covers the entire ML life cycle, giving students hands-on experience guided by world experts and developing their ML skills regardless of their background. The forums, chats, optional project, and online discussions with the class creators promote community development and continued learning. The early impact of this approach is demonstrable: numerous participants from a variety of locations and demographics have signed up. We have also begun initiatives to further increase access by helping develop courses that target K--12 students and teachers, as well as courses in other languages.
\section*{Acknowledgements}

Our approach to broadening access to applied machine learning using TinyML is based on input from many individuals at various organizations. We thank Rod Crawford, Tomas Edsö, and Felix Thomasmathibalan from \textbf{Arm}; Joe Lynas and Sacha Krstulović from \textbf{Audio Analytic}; Joshua Meyer from \textbf{Coqui}; Adam Benzio, Jenny Plunkett, Daniel Situnayake and Zach Shelby from \textbf{Edge Impulse}; Tulsee Doshi, Josh Gordon, Alex Gruenstein, Prateek Jain, and Nat Jeffries from \textbf{Google}; Marco Zennaro from \textbf{International Centre for Theoretical Physics (ICTP)}; Sek Chai from \textbf{Latent.AI}; Jane Polak Scowcroft from \textbf{Mozilla Common Voice}; Thiery Moreau from \textbf{OctoML}; Evgeni Gousev and Erich Plondke from \textbf{Qualcomm}; and Danilo Pau from \textbf{STMicroelectronics} for their valuable feedback. We are also grateful to the \textbf{Google TensorFlow Lite Micro team}, which includes Robert David, Jared Duke, Advait Jain, Vijay Janapa Reddi, Nat Jeffries, Jian Li, Nick Kreeger, Ian Nappier, Meghna Natraj, Shlomi Regev, Rocky Rhodes, and Tiezhen Wang, without whom we would have been unable to deploy models in microcontrollers, and \textbf{Arduino}---Jose Garcia Dotel and Martino Facchin---who helped us with global distribution of the TinyML kit. We also thank the tinyML Foundation for nurturing activity around embedded ML, providing guidance and supporting educational and outreach activities around TinyML.

\typeout{}

\bibliographystyle{unsrt}
\bibliography{references}

\end{document}